\documentclass[10pt,twocolumn,letterpaper]{article}

\usepackage[pagenumbers]{cvpr} 

\usepackage[dvipsnames]{xcolor}

\newcommand{\dlh}[1]{}
\newcommand{\tianfan}[1]{}
\newcommand{\myparagraph}[1]{\mbox{}\\[-7mm]\paragraph{#1}}
\newcommand{\model}{\textit{BiDiff}\xspace}

\definecolor{cvprblue}{rgb}{0.21,0.49,0.74}
\usepackage[pagebackref,breaklinks,colorlinks,citecolor=cvprblue]{hyperref}

\def\paperID{***} 
\def\confName{CVPR}
\def\confYear{2024}

\usepackage{hyperref}
\usepackage{url}
\usepackage[utf8]{inputenc} 
\usepackage[T1]{fontenc}    
\usepackage{hyperref}       
\usepackage[capitalize]{cleveref}
\usepackage{url}            
\usepackage{booktabs}       
\usepackage{amsfonts}       
\usepackage{nicefrac}       
\usepackage{microtype}      
\usepackage{xcolor}         
\newcommand{\bd}[1]{\boldsymbol{#1}}
\usepackage{graphicx}
\usepackage{wrapfig}

\title{Text-to-3D Generation with Bidirectional Diffusion using both 2D and 3D priors}

\author{%
  Lihe Ding$^{1,4}$\footnotemark[1], ~~Shaocong Dong$^{2}$\footnotemark[1], ~~Zhanpeng Huang$^{3}$, ~~Zibin Wang$^{3}$\footnotemark[2], \\
  Yiyuan Zhang$^{1}$, ~~Kaixiong Gong$^{1}$, ~~Dan Xu$^{2}$, ~~Tianfan Xue$^{1}$ \\
  {$^1$}The Chinese University of Hong Kong ~~{$^2$}Hong Kong University of Science and Technology \\
  {$^3$}SenseTime ~~{$^4$}Shanghai AI Laboratory \\
  {\tt\small\{dl023, gk023, tfxue\}@ie.cuhk.edu.hk, \{sdongae, danxu\}@cse.ust.hk}\\
  {\tt\small \{wangzb02, yiyuanzhang.ai\}@gmail.com, \{huangzhanpeng\}@sensetime.com}\\
}

\begin{document}


\twocolumn[{
	\renewcommand\twocolumn[1][]{#1}
	\maketitle
	\thispagestyle{empty}
	\vspace{-20pt}
	\begin{center}
            \includegraphics[width=0.9\linewidth]{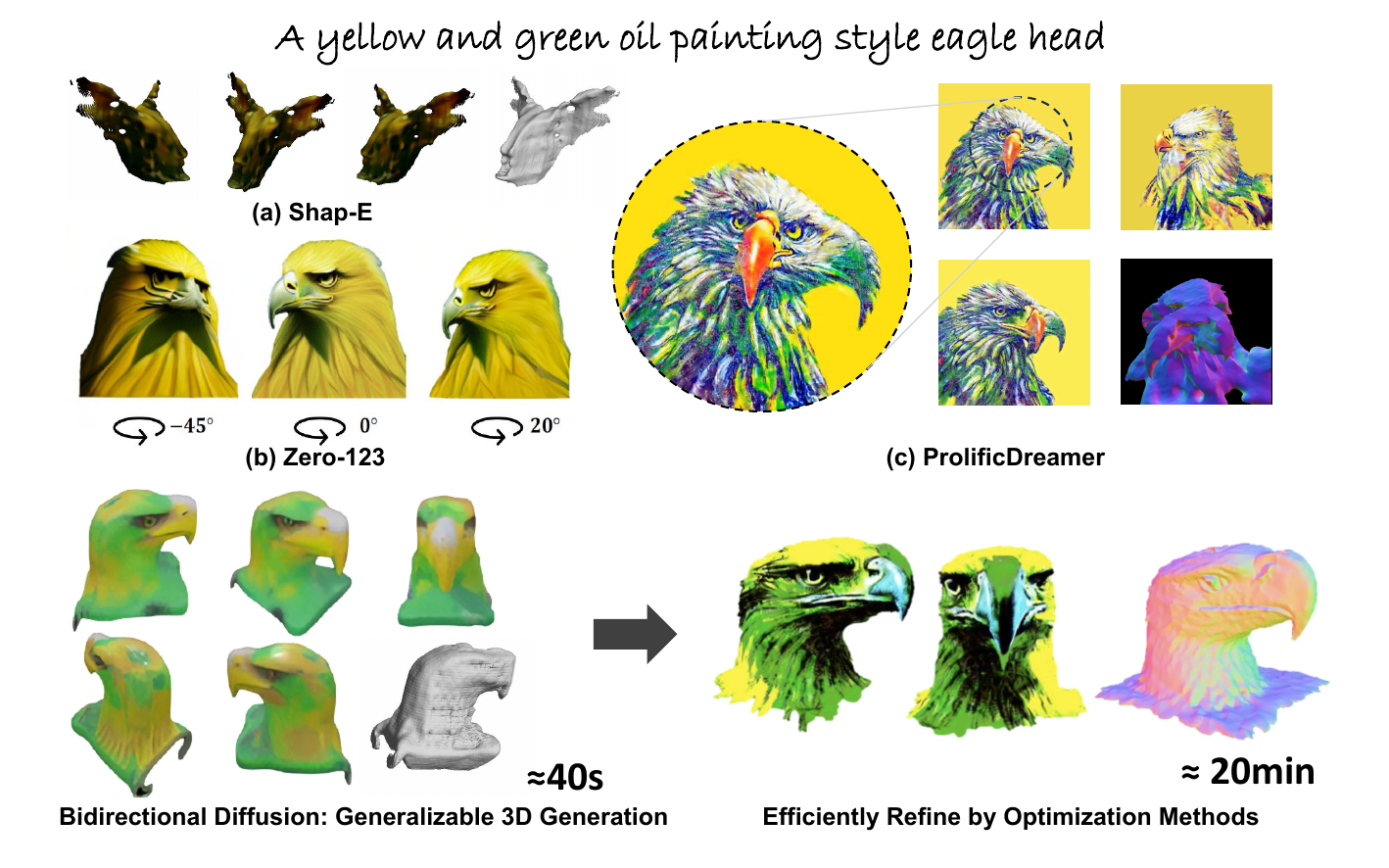}
             \vspace{-2pt}
            \captionof{figure}{
            Our \textit{BiDiff} can efficiently generate high-quality 3D objects. It alleviates all these issues in previous 3D generative models: (a) low-texture quality, (b) multi-view inconsistency, and (c) geometric incorrectness (e.g., multi-face Janus problem). The outputs of our model can be further combined with optimization-based methods (e.g., ProlificDreamer) to generate better 3D geometries with slightly longer processing time (bottom row).
            }
		\label{fig:teaser}
		\vspace{5pt}
	\end{center}
}]
\renewcommand{\thefootnote}{\fnsymbol{footnote}}
\footnotetext[1]{Equal contribution. Part of this work was done when Lihe Ding and Shaocong Dong interned at Sensetime.}
\footnotetext[2]{Corresponding author.}
\begin{abstract}
Most 3D generation research focuses on up-projecting 2D foundation models into the 3D space, either by minimizing 2D Score Distillation Sampling (SDS) loss or fine-tuning on multi-view datasets. Without explicit 3D priors, these methods often lead to geometric anomalies and multi-view inconsistency. Recently, researchers have attempted to improve the genuineness of 3D objects by directly training on 3D datasets, albeit at the cost of low-quality texture generation due to the limited texture diversity in 3D datasets. To harness the advantages of both approaches, we propose \textit{Bidirectional Diffusion} (\textit{BiDiff}), a unified framework that incorporates both a 3D and a 2D diffusion process, to preserve both 3D fidelity and 2D texture richness, respectively.
Moreover, as a simple combination may yield inconsistent generation results, we further bridge them with novel bidirectional guidance. 
In addition, our method can be used as an initialization of optimization-based models to further improve the quality of 3D model and efficiency of optimization, reducing the generation process from 3.4 hours to 20 minutes.
Experimental results have shown that our model achieves high-quality, diverse, and scalable 3D generation. Project website: \url{https://bidiff.github.io/}.
\end{abstract}    
\begin{figure*}[!t]
\setlength{\belowcaptionskip}{-0.3cm}
\centering{\includegraphics[width=0.98\linewidth]{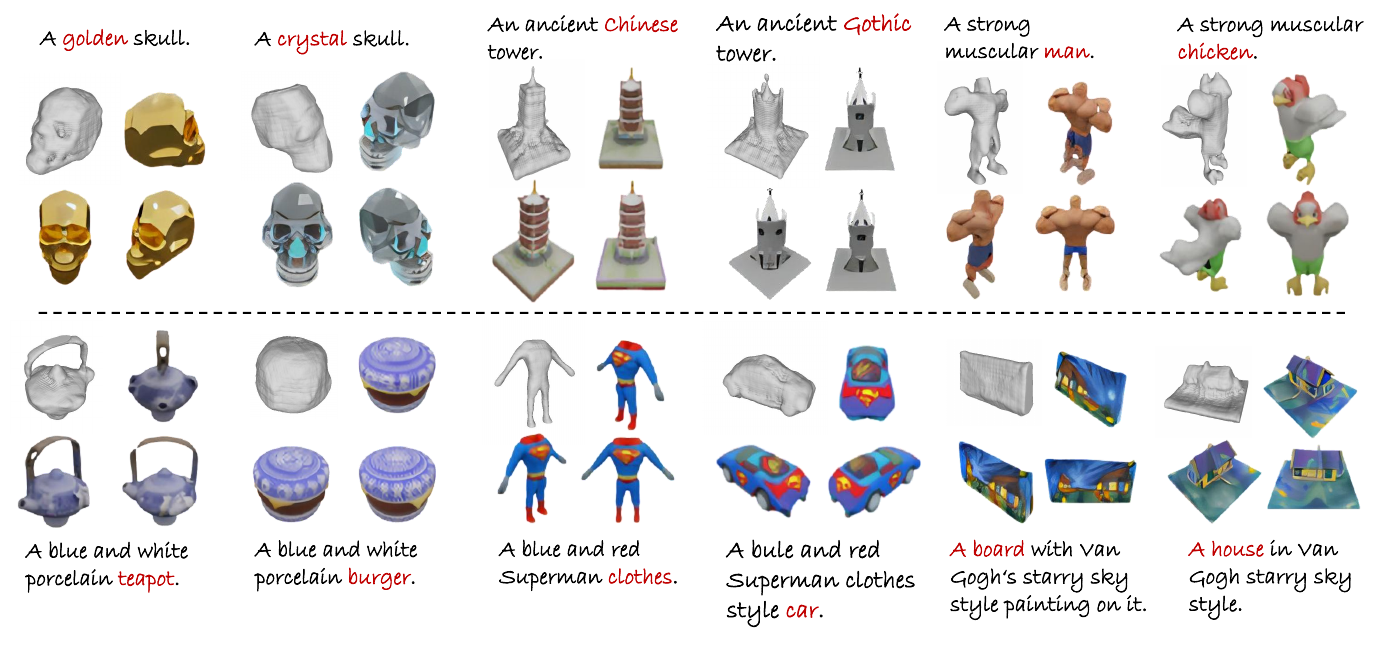}}
\vspace{-1em}
\caption{\textbf{Texture Control (Top)}: we change the texture while maintaining the overall shape. \textbf{Shape Control (Bottom)}: we fix texture patterns and generate various shapes.}
\label{fig:text_control}
\vspace{-0.5em}
\end{figure*}

\section{Introduction}
\label{sec:intro}

Recent advancements in text-to-3D generation~\citep{metzer2022latent} mainly focus on lifting 2D foundation models into 3D space. One of the most popular solutions~\citep{poole2022dreamfusion, lin2022magic3d} uses 2D Score Distillation Sampling (SDS) loss derived from a 2D diffusion model to supervise 3D generation. While these methods can generate high-quality textures, they often lead to geometric ambiguity, such as the multi-face Janus problem~\cite{metzer2023latent}, due to the lack of 3D constraints (\cref{fig:teaser}(c)). Moreover, these optimization methods are time-consuming, taking hours to generate one object. Zero-123\cite{liu2023zero} tries to alleviate the problem by fine-tuning the 2D diffusion models on multi-view datasets, but it still cannot guarantee geometric consistency (\cref{fig:teaser}(b)).

To ensure better 3D consistency, another solution is to directly learn 3D structures from 3D datasets~\citep{nichol2022point, jun2023shap}. However, many existing 3D datasets~\citep{chang2015shapenet, deitke2022objaverse} only contain handcrafted objects or lack high-quality 3D geometries, with textures very different from real-world objects. Moreover, 3D datasets are often much smaller than, and also difficult to scale up to, their 2D counterparts. As a result, the 3D diffusion models (\cref{fig:teaser} (a)) normally cannot generate detailed textures and complicated geometry, even if they have better 3D consistency compared to up-projecting 2D diffusion models.

Therefore, a straightforward way to leverage the advantages of both methods is to combine both 2D and 3D diffusion models. However, a simple combination may result in inconsistent generative directions as they are learned in two independent diffusion processes. In addition, the two diffusion models are represented in separate 2D and 3D spaces without knowledge sharing.

To overcome these problems, we propose \textit{Bidirectional Diffusion} (\model), a method to seamlessly integrate both 2D and 3D diffusion models within a unified framework. Specifically, we employ a hybrid representation in which a signed distance field (SDF) is used for 3D feature learning and multi-view images for 2D feature learning. The two representations are mutually transformable by rendering 3D feature volume into 2D features and back-projecting 2D features to 3D feature volume. Starting from pretrained 3D and 2D diffusion models, the two diffusion models are jointly finetuned to capture a joint 2D and 3D prior facilitating 3D generation.

However, correlating the 2D and 3D representations is not enough to combine two diffusion processes, as they may deviate from each other in the following diffusion steps. To solve this problem, we further introduce bidirectional guidance to align the generative directions of the two diffusion models. At each diffusion step, the intermediate results from the 3D diffusion scheme are rendered into 2D images as guidance signals to the 2D diffusion model. Meanwhile, the multi-view intermediate results from the 2D diffusion process are also back-projected to 3D, guiding the 3D diffusion. The mutual guidance regularizes the two diffusion processes to learn in the same direction.

The proposed bidirectional diffusion poses several advantages over the previous 3D generation models. First, users can separately control the generation of 2D texture and 3D geometry, as shown in \cref{fig:text_control}, because the 2D diffusion model focuses on texture generation and the 3D diffusion model focuses on geometry. This is impossible for previous 3D diffusion methods.
Secondly, compared to 3D-only diffusion models~\citep{jun2023shap}, our method takes advantage of a 2D diffusion model trained on much larger datasets. Therefore, it can generate more diversified objects and create a completely new object like ``A strong muscular chicken'' illustrated in Fig~\ref{fig:text_control}.
Thirdly, compared to previous optimization methods~\citep{poole2022dreamfusion,wang2023prolificdreamer} that often take several hours to generate one object, we utilize a fast feed-forward joint 2D-3D diffusion model for scalable generation, which only takes about 40 seconds to generate one object.

Moreover, because of the efficacy of \model, we also propose an optional step to utilize its output as an initialization for the existing optimization-based methods (e.g., ProlificDreamer~\cite{wang2023prolificdreamer}). This optional step can further improve the quality of a 3D object, as demonstrated in the bottom row of \cref{fig:teaser}. Also, the good initialization from \model helps to reduce optimization time from around 3.4 hours to 20 minutes, and concurrently resolves geometrical inaccuracy issues, like multi-face anomalies. Moreover, this two-step generation enables creators to rapidly adjust prompts to obtain a satisfactory preliminary 3D model through a lightweight feed-forward generation process, subsequently refining it into high-fidelity results.

Through training on ShapeNet~\citep{chang2015shapenet} and Objaverse 40K~\citep{deitke2022objaverse}, our framework is shown to generate high-quality textured 3D objects with strong generalizability. In summary, our contributions are as follows: 1) We propose \textit{BiDiff}, a joint 2D-3D diffusion model, that can generate high-quality, 3D-consistent, and diversified 3D objects; 2) We propose a novel training pipeline that utilizes both pretrained 2D and 3D generative foundation models; 3) We propose the first diffusion-based 3D generation model that allows independent control of texture and geometry; 4) We utilize the outputs from \textit{BiDiff} as a strong initialization for the optimization-based methods, generating high-quality geometries while ensuring that users receive quick feedback for each prompt update.
\vspace{-.3em}
\section{Related Work}
\label{gen_inst} 
\vspace{-0.3em}
Early 3D generative methods adopt various 3D representations, including 3D voxels~\citep{Wu2016,Edward2017,Henzler2019}, point clouds~\citep{Achlioptas2018,yang2019pointflow}, meshes~\citep{Gao2019,Ibing2021}, and implicit functions~\citep{Chen2019,Park2019} for category-level 3D generations. These methods directly train the generative model on a small-scale 3D dataset, and, as a result, the generated objects may either miss tiny geometric structures or lose diversity. Even though there are large-scale~\citep{deitke2022objaverse} or high-quality 3D datasets~\citep{wu2023omniobject3d} in recent years, they are still much smaller than the datasets used for 2D image generation training.  

With the powerful text-to-image synthesis models~\citep{radford2021clip,saharia2022photorealistic,rombach2022high}, a new paradigm emerges for 3D generation without large-scale 3D datasets by leveraging 2D generative model.
One line of works utilizes 2D priors from pre-trained text-to-image model (known as CLIP)~\citep{jain2022zero,khalid2022clipmesh} or 2D diffusion generative models \citep{wang2022score,lin2022magic3d,metzer2022latent} to guide the optimization of underlying 3D representations.
However, these models could not guarantee cross-view 3D consistency and the per-instance optimization scheme suffers both high computational cost and over-saturated problems.
Later on, researchers improve these models using textual codes or depth maps~\citep{seo2023let,deng2023nerdi,melaskyriazi2023realfusion}, and 
\cite{wang2023prolificdreamer} directly model 3D distribution to improve diversity. These methods alleviate the visual artifacts but still cannot guarantee high-quality 3D results.

Another line of works learn 3D priors directly from 3D datasets. As the diffusion model has been the de-facto network backbone for most recent generative models, it has been adapted to learn 3D priors using implicit spaces such as point cloud features~\citep{zeng2022lion,nichol2022point}, NeRF parameters~\citep{jun2023shap,erkoç2023hyperdiffusion}, or SDF spaces ~\citep{cheng2022sdfusion,Liu2023MeshDiffusion}. The synthesized multi-view images rendered from 3D datasets were also utilized to provide cross-view 3D consistent knowledge ~\citep{liu2023zero}. These methods normally highlight fast inference and 3D consistent results. However, due to inferior 3D dataset quality and size, these methods generally yield visually lower-quality results with limited diversity. Recently a few methods~\citep{qian2023magic123,shi2023MVDream} explored to combine 2D priors and 3D priors from individual pre-trained diffusion models, but they often suffer from inconsistent between two generative processes.
\begin{figure*}[tbp]
\setlength{\belowcaptionskip}{-0.3cm}
\centering{\includegraphics[width=0.98\linewidth]{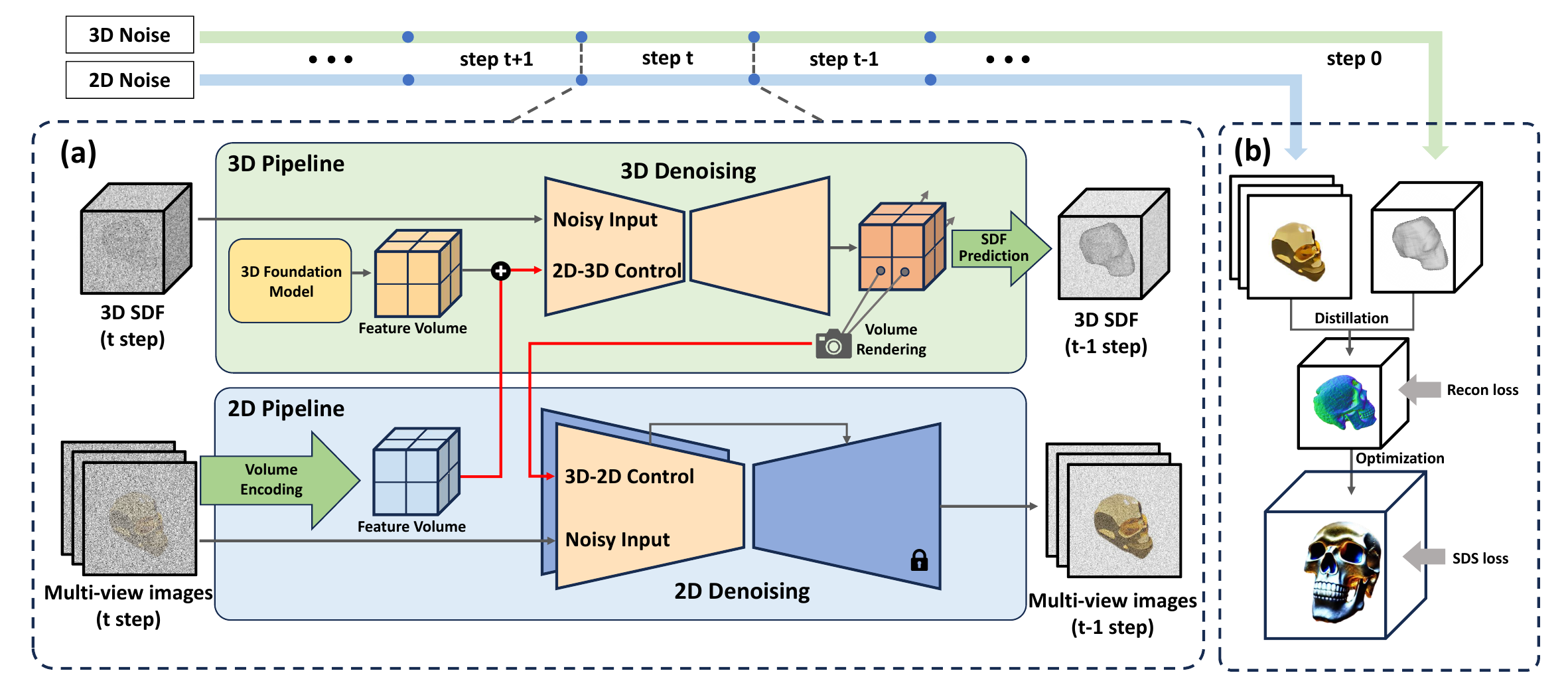}}
\vspace{-1em}
\caption{The \textit{BiDiff} framework operates as follows: (a) At each step of diffusion, we render the 3D diffusion's intermediate outputs into 2D images, which then guide the denoising of the 2D diffusion model. Simultaneously, the intermediate multi-view outputs from the 2D diffusion are re-projected to assist the denoising of the 3D diffusion model. Red arrows show the bidirectional guidance, which ensures that both diffusion processes evolve coherently. (b) We use the outcomes of the 2D-3D diffusion as a strong starting point for optimization methods, allowing for further refinement with fewer optimization steps.}
\label{fig:arch}
\vspace{-.3em}
\end{figure*}

\vspace{-.3em}
\section{Method}
\label{headings}
\vspace{-.3em}
As many previous studies~\cite{liu2023zero,qian2023magic123} have illustrated, both 2D texture and 3D geometry are important for 3D object generation. However, incorporating 3D structural priors and 2D textural priors is challenging: i) combining both 3D and 2D generative models into a single cohesive framework is not trivial;  ii) in both training and inference, two generative models may lead to opposite generative directions. 

To tackle these problems, we propose \model, a novel bidirectional diffusion model that marries a pretrained 3D diffusion model with another 2D one using bidirectional guidance. \cref{fig:arch} illustrates the overall architecture of our framework. Details of each component will be discussed below. Specifically, in \cref{sec:bidiff}, we will introduce our novel hybrid representation that includes both 2D and 3D information, and the bidirectional diffusion model built on top of this hybrid representation. In \cref{sec:3d_diff_with_2d_guide} and \cref{sec:2d_diff_with_3d_guide}, to ensure the two generative models lead to the same generative direction, we will introduce how to add bidirectional guidance to both 3D and 2D diffusion models. In \cref{sec:prior_enhance}, we discuss one advantage of \model, which is independent control of texture and geometry generation, as shown in ~\cref{fig:text_control}. Finally, in \cref{sec:post_optim}, we discuss another advantage of \model, which is to use the results from \model as a strong initialization for optimization-based methods to obtain more delicate results efficiently.

\subsection{Bidirectional Diffusion}
\label{sec:bidiff}
\vspace{-.3em}
To incorporate both 2D and 3D priors, we represent a 3D object using a hybrid combination of two formats: Signed Distance Field (SDF) $\mathcal{F}$ and multi-view image set $\mathcal{V} = \left\{ \mathcal{I}^i \right\}_{i=1}^{M}$, where $\mathcal{F}$ is computed from signed distance values on an $N \times N \times N$ grid, and $I^{i}$ is the $i$-th image from a multi-view image set of size $M$. This hybrid representation is shown on the left side of \cref{fig:arch}.

With this representation, we learn a joint distribution \( \left \{ \mathcal{F}, \mathcal{V} \right \} \) utilizing two distinct diffusion models: a 3D diffusion model \( \mathcal{D}_{3d} \) in the SDF space (the green 3D denoising block in \cref{fig:arch}) and a 2D multi-view diffusion model \( \mathcal{D}_{2d} \) within the image domain (the blue 2D denoising block in \cref{fig:arch}).
Specifically, given a timestep $t$, we add Gaussian noises to both SDF and multi-view images as 
\vspace{-1mm}
\begin{equation}
\begin{split}
    \mathcal{F}_{t} = \sqrt{\overline{\alpha}_{t}} \mathcal{F}_{0} + \sqrt{1 - \overline{\alpha}_{t}} \epsilon_{3d}
    \text{~~and~~} \\ 
    \mathcal{I}_{t}^{i} = \sqrt{\overline{\alpha}_{t}}
    \mathcal{I}_{0}^{i} + \sqrt{1 - \overline{\alpha}_{t}} \epsilon^{i}_{2d}\text{~for~}\forall i,
\end{split}
\end{equation}
where $\epsilon \sim \mathcal{N}(0, \textbf{I})$ is random noise, and $\overline{\alpha}_t$ is a noise schedule which is different in 3D and 2D. Subsequently, the straightforward way is to separately train these two diffusion models by minimizing the following two objectives:
\vspace{-1mm}
\begin{equation}
\begin{split}
    L_{simple3d} &= \mathbb{E}_{\mathcal{F}_{t}, \epsilon_{3d}, t}\| \epsilon_{3d} - \mathcal{D}_{3d}(\mathcal{F}_t, t)\|_{2}^{2},\\ 
    L_{simple2d} &= \frac{1}{N} \sum_{i=1}^{N}( \mathbb{E}_{\mathcal{I}_{t}^{i}, \epsilon^{i}_{2d}, t }\| \epsilon^{i}_{2d} - \mathcal{D}_{2d}(\mathcal{I}_t^{i}, t)\|_{2}^{2} ),
\end{split}
\end{equation}
where $\epsilon_{3d}$ and $\epsilon_{2d}$ are Gaussian noises $\epsilon_{3d},\epsilon^{i}_{2d} \sim \mathcal{N}(0, \textbf{I})$, SDF and image set are sampled from forward diffusion processes $\mathcal{F}_{t} \sim q(\mathcal{F}_{t}), \mathcal{I}_{t}^{i} \sim q(\mathcal{I}_{t}^{i})$, and timestep is uniformly sampled $t \sim U[1, T]$.

However, this simple combination does not consider the correlations between 3D and 2D diffusion, which may hinder the understanding of 2D and 3D consistency, leading to inconsistent generation between 3D geometry and 2D multi-view images.

\tianfan{if running out of space, the following few paragraphs can be deleted.}

We resolve this problem by a novel \textit{Bidirectional Diffusion}. In this model, the consistency between 3D and 2D diffusion output is enforced through bidirectional guidance. First, we add guidance from the 2D diffusion process to the 3D generative process, which is the red arrow pointing to the ``2D-3D control''. Specifically, during each denoising step $t$, we feed the denoised multi-view images $\mathcal{V}_{t+1}' = \left \{ \mathcal{I}_{t+1}^i \right \}_{i=1}^{N}$ in previous step into the 3D diffusion model as $\epsilon'_{3d} = \mathcal{D}_{3d}(\mathcal{F}_{t}, \mathcal{V}_{t+1}', t)$. This guidance steers the current 3D denoising direction to ensure 2D-3D consistency.
It's worth mentioning that the denoised output $\mathcal{V}_{t+1}'$ from the previous step $t+1$ is inaccessible in training, therefore we directly substitute it with the ground truth $\mathcal{V}_{t}$. In inference, we utilize the denoised images from the previous step. Then we could obtain the denoised radiance field $\mathcal{F}_{0}'$ given the 2D guided noise prediction $\epsilon_{3d}'$ by $\mathcal{F}_{0}' = \frac{1}{\sqrt{\overline{\alpha}_{t}}}(\mathcal{F}_{t} - \sqrt{1 - \overline{\alpha}_{t}} \epsilon'_{3d}).$

Secondly, we also add guidance from the 3D diffusion process to the 2D generative process. Specifically, using the same camera poses, we render multi-view images $\mathcal{H}_{t}^{i}$ derived from the radiance field $\mathcal{F}_{0}'$ by the 3D diffusion model: $\mathcal{H}_{t}^{i} = \mathcal{R}(\mathcal{F}_{0}', \mathcal{P}^{i}), i=1,...M$, where $\mathcal{P}^{i}$ is the $i^{th}$ camera pose. These images are further used as guidance to the 2D multi-view denoising process $\mathcal{D}_{2d}$ by $\epsilon'_{2d} = \mathcal{D}_{2d}(\mathcal{V}_{t}, \left \{ \mathcal{H}_{t}^{i} \right \}_{i=1}^{N}, t).$. This guidance is the red arrow pointing to the ``3D-2D control'' in \cref{fig:arch}.

Our method can seamlessly integrate and synchronize both the 3D and 2D diffusion processes within a unified framework. In the following sections, we will delve into each component in detail.

\subsection{3D Diffusion Model with 2D Guidance}
\label{sec:3d_diff_with_2d_guide}
\vspace{-.3em}
Our 3D diffusion model aims to generate a neural surface field (NeuS) ~\cite{long2022sparseneus}, with novel 2D-to-3D guidance derived from the denoised 2D multi-view images.
To train our 3D diffusion model, at each training timestep $t$, we add noise to a clean radiance field, yielding a noisy one $\mathcal{F}_{t}$. This field, combined with the timestep $t$ embeddings and the text embeddings, is then passed through 3D sparse convolutions to generate a 3D feature volume $\mathcal{M}$ as: $\mathcal{M} = \text{Sp3DConv}(\mathcal{F}_{t}, t, \text{text}).$
Then we sample $N \times N \times N$ grid points from $\mathcal{M}$ and project these points onto all denoised multi-view images $\mathcal{V}_{t+1}'$ from the previous step of the 2D diffusion model. At each grid point $p$, we aggregate the interpolated 2D feature at its 2D projected location on each view, and calculate the mean and variance over all $N$ interpolated features to obtain the image-conditioned feature volume $\mathcal{N}$:
\begin{equation}
    \mathcal{N}(p) = [\text{Mean}(\mathcal{V}_{t+1}'(\pi(p))), \text{Var}(\mathcal{V}_{t+1}'(\pi(p)))],
\end{equation}
where $\pi$ denotes the projection operation from 3D to 2D image plane. We fuse these two feature volumes with further sparse convolutions for predicting the clean $\mathcal{F}_0$.

One important design of our 3D diffusion model is that it incorporates geometry priors derived from the 3D foundation model, Shap-E~\citep{jun2023shap}. Shap-E is a latent diffusion~\citep{metzer2022latent} model trained on several millions 3D objects, and thus ensures the genuineness of generated 3D objects. 
Still, we do not want Shap-E to limit the creativity of our 3D generative model, and try to preserve the capability of generating novel objects that Shap-E cannot.

To achieve this target, we design a feature volume $\mathcal{G}$ to represent a radiance field converted from the Shap-E latent code $\mathcal{C}$. It is implemented using NeRF MLPs by setting their parameters to the latent code $\mathcal{C}$: $\mathcal{G}(p) = \text{MLP}(\lambda(p); \theta=\mathcal{C}),$ where $\lambda$ denotes the positional encoding operation.

One limitation of directly introducing Shap-E latent code is that the network is prone to shortcut the training process, effectively memorizing the radiance field derived from Shap-E. To generate 3D objects beyond Shap-E model, we add Gaussian noise at level $t_0$ to the clean latent code, resulting in the noisy latent representation $\mathcal{C}_{t_0}$, where $t_0$ represents a predefined constant timestep. Subsequently, the noisy radiance field $\mathcal{G}_{t_0}$ is decoded by substituting $\mathcal{C}$ with $\mathcal{C}_{t_0}$. This design establishes a coarse-to-fine relationship between the 3D prior and the ground truth, prompting the 3D diffusion process to leverage the 3D prior without excessively depending on it.

In this way, we can get the fused feature volume as:
\begin{equation}
    \mathcal{S} = \mathcal{U}([\mathcal{M}, \text{Sp3DConv}(\mathcal{N}), \text{Sp3DConv}(\mathcal{G}_{t_0})]),
\end{equation}
where $\mathcal{U}$ denotes 3D sparse U-Net. Then we can query features from $\mathcal{S}$ for each grid point $p$ and decode it to SDF values through several MLPs: $\mathcal{F}'_{0}(p) = \text{MLP}(\mathcal{S}(p), \lambda(p)),$
where $\mathcal{S}(p)$ represents the interpolated features from $\mathcal{S}$ at position $p$. In \cref{sec:compare} and \cref{fig:qualitative}, our experiments also demonstrate that our model can generate 3D objects beyond Shap-E model.

\subsection{2D Diffusion Model with 3D Guidance}
\label{sec:2d_diff_with_3d_guide}
\vspace{-.3em}
Our 2D diffusion model simultaneously generates multi-view images by jointly denoising multi-view noisy images $\mathcal{V}_t = \left \{ \mathcal{I}_t^i \right \}_{i=1}^{M}$.
To encourage 2D-3D consistency, the 2D diffusion model is also guided by the 3D radiance field output from 3D diffusion process mentioned above. Specifically, for better image quality, 2D multi-view diffusion model is built on the multiple independently frozen 2D foundation models (e.g., DeepFloyd~\cite{deepfloyd_if}) to harness the potent 2D priors. Each of these frozen 2D foundation models (the dark blue network in \cref{fig:arch}) is modulated by view-specific 3D-consistent residual features and responsible for the denoising of a specific view, as described below.

First, to achieve 3D-to-2D guidance, we render multi-view images from the 3D denoised radiance field $\mathcal{F}'_{0}$ and feed them to 2D denoising model. Note that the radiance field consists of a density field and a color field. The density field is constructed from the signed distance field (SDF) generated by our 3D diffusion model using S-density introduced in NeuS~\citep{wang2021neus}. To obtain the color field, we apply another color MLP to the feature volume in the 3D diffusion process.

Upon obtaining the color field $c$ and density field $\sigma$, we conduct volumetric rendering on each ray $\bd{r}(m) = \bd{o} + m\bd{d}$ which extends from the camera origin $\bd{o}$ along a direction $\bd{d}$ to produce multi-view consistent images $\left \{ \mathcal{H}^{i} \right \}_{i=1}^{M}$:
\begin{equation}
    \hat{C}(\bd{r}) = \int_{0}^{\infty}T(m)\sigma(\bd{r}(m)))c(\bd{r}(m)), \bd{d})dm,
\end{equation}
where $T(m) = \rm{exp}(-\int_0^{m}\sigma(\bd{r}(s))ds)$ handles occlusion.

Secondly, we use these rendered multi-view images as guidance for the 2D foundation model. We first use a shared feature extractor $\mathcal{E}$ to extract hierarchical multi-view consistent features from these images. Then each extracted feature is added as residuals to the decoder of its corresponding frozen 2D foundation denoising U-Net (the red arrow pointing to ``3D-2D Control'' in \cref{fig:arch}), achieving multi-view modulation and joint denoising following ControlNet~\citep{zhang2023controlnet} as
$ \hat{\bd{f}_{k}^{i}} = \bd{f}_{k}^{i} + \text{ZeroConv}(\mathcal{E}(\mathcal{H}^{i})[k]),$
where $\bd{f}_i^k$ denotes the original feature maps of the $k$-th decoder layer in 2D foundation model, $\mathcal{E}(\mathcal{H}^{i})[k]$ denotes the $k$-th residual features of the $i$-th view, and ZeroConv~\citep{zhang2023controlnet} is $1 \times 1$ convolution which is initialized by zeros and gradually updated during training. Experimental results show that this 3D-to-2D guidance helps to ensure multi-view consistency and facilitate geometry understanding.

\subsection{Separate Control of Geometry and Texture}
\vspace{-.3em}
\label{sec:prior_enhance}
One advantage of \model is that it naturally separates 2D texture generation using 2D diffusion model from 3D geometry generation using 3D diffusion model. Because of this, users can separately control geometry and texture generation, as shown in \cref{fig:text_control}.

To achieve this, we first propose a prior enhancement strategy to empower a manual control of the strength of 3D and 2D priors independently.
Inspired by the classifier-free guidance~\citep{ho2022classifier}, during training, we randomly drop the information from 3D priors by setting condition feature volume from $\mathcal{G}$ to zero and weaken the 2D priors by using empty text prompts.
Consequently, upon completing the training, we can employ two guidance scales, $\gamma_{3d}$ and $\gamma_{2d}$, to independently modulate the influence of these two priors.

Specifically, to adjust the strength of 3D prior, we calculate the difference between 3D diffusion outputs with and without conditional 3D feature volumes, and add them back to 3D diffusion output:
\begin{equation}
\begin{split}
\label{eq:3d_guide}
    \hat{\epsilon}_{3d} = & \mathcal{D}_{3d}(\mathcal{F}_t, \mathcal{V}_{t+1}', t) + \gamma_{3d} \cdot ((\mathcal{D}_{3d}(\mathcal{F}_t, \mathcal{V}_{t+1}', t| \mathcal{G}) - \\
    &\mathcal{D}_{3d}(\mathcal{F}_t, \mathcal{V}_{t+1}', t)).
\end{split}
\end{equation}
Then we can control the strength of 3D prior by adjusting the weight $\gamma_{3d}$ of this difference term. When $\gamma_{3d} = 0$, it will completely ignore 3D prior. When $\gamma_{3d} = 1$, this is just the previous model that uses both 3D prior and 2D prior. When $\gamma_{3d} > 1$, the model will produce geometries close to the conditional radiance field but with less diversity.

Also, we can similarly adjust the strength of 2D priors by adding differences between 2D diffusion outputs with and without conditional 2D text input:
\begin{equation}
\begin{split}
\label{eq:2d_guide}
    \hat{\epsilon}_{2d} = &\mathcal{D}_{2d}(\mathcal{V}_t, \left \{ \mathcal{H}_{t}^{i} \right \}_{i=1}^{M}, t) + \\
    &\gamma_{2d} \cdot ((\mathcal{D}_{2d}(\mathcal{V}_t, \left \{ \mathcal{H}_{t}^{i} \right \}_{i=1}^{M}, t| text)) - \\
    &\mathcal{D}_{2d}(\mathcal{V}_t, \left \{ \mathcal{H}_{t}^{i} \right \}_{i=1}^{M}, t)).
\end{split}
\end{equation}
Increasing $\gamma_{2d}$ results in more coherent textures with text, albeit at the expense of diversity.
It is worth noting that while we adjust the 3D and 2D priors independently via~\cref{eq:3d_guide} and~\cref{eq:2d_guide}, the influence inherently propagates to the other domain due to the intertwined nature of our bidirectional diffusion process.

With these two guidance scales $\gamma_{3d}$ and $\gamma_{2d}$, we can easily achieve a separate control of geometry and texture. First, to only change texture while keep geometry untouched, we just fix the initial 3D noisy SDF grids and the conditional radiance field $\mathcal{C}_{t_0}$, while enlarge its influence by \cref{eq:2d_guide}. On the other hand, to only change geometry while keep texture style untouched, we can maintain keywords in text prompts and enlarge its influence by \cref{eq:3d_guide}. By doing so, the shape will be adjusted by the 3D diffusion process.

\subsection{Optimization with BiDiff Initialization}
\vspace{-.3em}
\label{sec:post_optim}
The generated radiance field $\mathcal{F}_0$ using \model can be further used as a strong initialization of the optimization-based methods~\cite{wang2023prolificdreamer}. This additional step can further improve the quality of the 3D model, as shown in \cref{fig:teaser} and \cref{fig:comp_all}. Importantly, compared to the geometries directly generated by optimization, our \model can output more diversified geometry and generated geometries better aligns with users' input text, and also has more accurate 3D geometry. Therefore, the optimization started from this strong initialization can be rather efficient ($\approx$ 20min) and avoid incorrect geometries like multi-face and floaters.

Specifically, we first convert generated radiance field $\mathcal{F}_0$ from \model into a higher resolution one $\overline{\mathcal{F}}_0$ that supports $512 \times 512$ resolution image rendering, as shown on the right of \cref{fig:arch}. This process is achieved by a fast NeRF distillation operation ($\approx $ 2min). The distillation first bounds the occupancy grids of $\overline{\mathcal{F}}_0$ with the estimated binary grids (transmittance $ > 0.01$) from the original radiance field $\mathcal{F}_0$, then overfits $\overline{\mathcal{F}}_0$ to $\mathcal{F}_0$ by minimizing both the $L_1$ distance between two density fields and $L_1$ distance between their renderings 2D images under random viewpoints. Thanks to this flexible and fast distillation operation, we can efficiently convert generated radiance field from \model into any 3D representations an optimization-based method requires. In our experiments, since we are using ProlificDreamer~\cite{wang2023prolificdreamer}, we use the InstantNGP~\cite{mueller2022instant} as the high-resolution radiance field.

After initialization, we optimize $\overline{\mathcal{F}}_0$ by SDS loss following the previous methods~\cite{poole2022dreamfusion, wang2023prolificdreamer}. It is noteworthy that since we already have a good initialized radiance field, we only need to apply a small noise level SDS loss. Specifically, we set the ratio range of denoise timestep $t_{opt}$ to [0.02, 0.5] during the entire optimization process. \tianfan{Low-priority: denoise timestep is less informative. Better to mention diffusion steps, and compare that with the pure optimization method.}
\section{Experiment}

\begin{figure*}[htbp]
\centering{\includegraphics[width=0.98\textwidth]{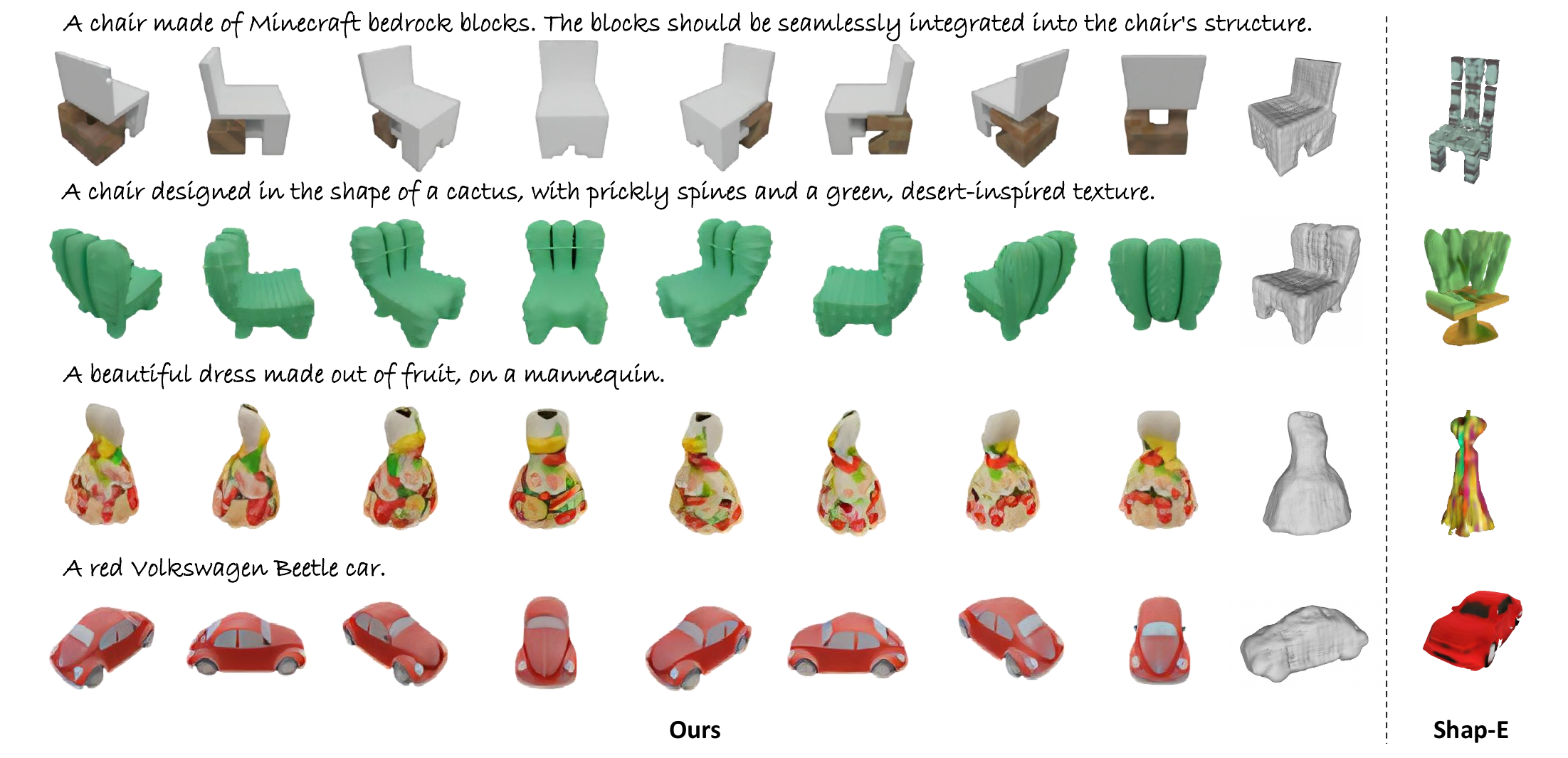}}
\caption{Qualitative sampling results of Bidirectional Diffusion model, including multi-view images and 3D mesh from diffusion sampling. The top two rows are the results on the Shapenet-Chair, and the bottom two rows are the results on the Objaverse. We compared the results of Shap-E in the last column.}
\vspace{-1em}
\label{fig:qualitative}
\end{figure*}

In this section, we described our experimental results. We train our framework on the ShapeNet-Chair~\citep{chang2015shapenet} and Objaverse LVIS 40k datasets~\citep{deitke2022objaverse}. We use the pre-trained DeepFloyd-IF-XL~\cite{deepfloyd_if} as our 2D foundation model and Shap-E~\citep{jun2023shap} as our 3D priors. We adopt the SparseNeuS~\citep{long2022sparseneus} as the neural surface field presentation with $N=128$. For the 3D-to-2D guidance, We follow the setup of ControlNet~\citep{zhang2023controlnet} to render $M=8$ multi-view images with $64 \times 64$ resolution using SparseNeuS. We train our framework on 4 NVIDIA A100 GPUs for both ShapeNet and Objaverse 40k experiments with batch size of 4. During sampling, we set the 3D and 2D prior guidance scale to 3.0 and 7.5 respectively. More details on data processing and model architecture are included in supplementary material. We discuss the evaluation and ablation study results below. Also, \textbf{please refer to supplementary webpages and videos for more visual results}.

\subsection{Text-to-3D Results}
\paragraph{ShapeNet-Chair results.}
The first and second rows of \cref{fig:qualitative} present our results trained on the ShapeNet-Chair dataset. Although the chair category often contains complicated geometric details, our framework demonstrates the capability to capture those fine details. Concurrently, our approach exhibits a remarkable capability to produce rich and diverse textures by merely modulating the textual prompts, leading to compelling visual outcomes.

\vspace{-.7em}
\paragraph{Objaverse-40K results.}
Scaling to a much larger 3D dataset, Objaverse-40K, our framework's efficacy becomes increasingly pronounced. The bottom two rows of \cref{fig:qualitative} are results from the Objaverse dataset. Compared to objects generated by Shap-E, our model closely adheres to the given textual prompts. This again shows that the proposed \model learns to model both 2D textures and 3D geometries better compared with 3D-only solutions, and is capable of generating more diverse geometries. 
\paragraph{Decouple geometry and texture control.}
\begin{wraptable}{r}[0.4cm]{4.5cm}
    \setlength{\abovecaptionskip}{-0.5mm}  
    \vspace{-4mm}
    \caption{CLIP R-precision.}
    \label{table:clipr}
    \centering
    \small
    \hspace{-6mm}
    \begin{tabular}
    {c@{\hspace{0.12cm}}c@{\hspace{0.12cm}}c}
        \toprule
        Method & R-P & time  \\
        \midrule
        DreamFusion & 0.67 & 1.1h \\
        ProlificDreamer & 0.83 & 3.4h  \\
        Ours-sampling & 0.79 & \textbf{40s}  \\
        Ours-post & \textbf{0.85} & 20min  \\
        \bottomrule
    \end{tabular}
    \vspace{-4mm}
    \label{tab:rp}
\end{wraptable}
Lastly, we illustrate that our \model can separately control geometry and texture generation. First, as illustrated in the first row of \cref{fig:text_control}, when the 3D prior is fixed, we have the flexibility to manipulate the 2D diffusion model using varying textual prompts to guide the texture generation process. This capability enables the generation of a diverse range of textured objects while maintaining a consistent overall shape.
Second, when we fix the textual prompt for the 2D priors (e.g., "a xxx with Van Gogh starry sky style"), we can adjust the 3D diffusion model by varying the conditional radiance field derived from the 3D priors. This procedure results in the generation of a variety of shapes, while maintaining a similar texture, as shown in the second row of \cref{fig:text_control}.

\begin{figure*}[htbp]
\centering{\includegraphics[width=0.95\textwidth]{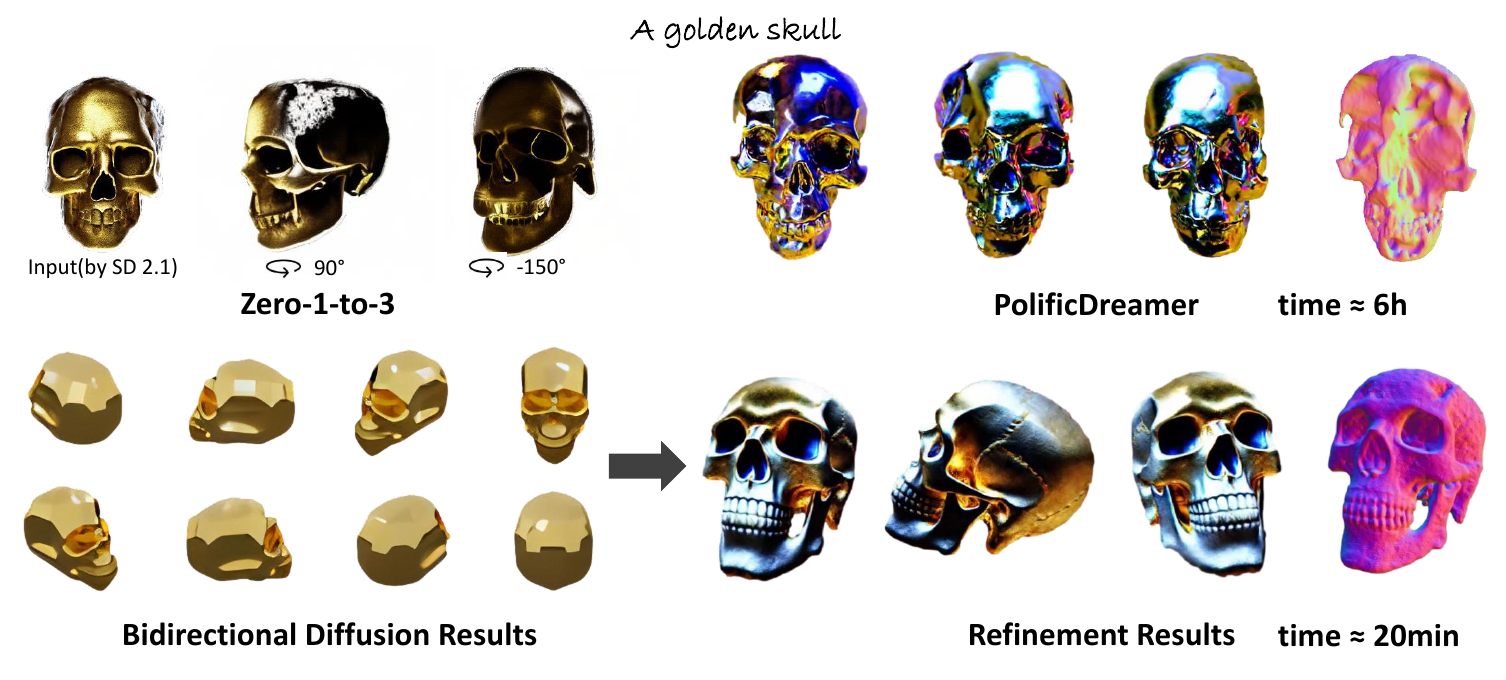}}
\caption{Comparison with other optimization or multi-view diffusion based works. We show both multi-view images (left) and 3D results (right). Zero-1-to-3~\cite{liu2023zero} is not good at predicting results from a large perspective, and PolificDreamer~\cite{wang2023prolificdreamer} suffers from the multi face problem. Our method has excellent robustness and can obtain high-quality results in a short period of time.}
\label{fig:comp_all}
\vspace{-1em}
\end{figure*}

\subsection{Comparison with other Generation Models}
\label{sec:compare}
\vspace{-1em}

\vspace{-1em}
\myparagraph{Comparison with optimization methods.}
Our framework is capable of simultaneously generating multi-view consistent images alongside a 3D mesh in a scalable manner. In contrast, the SDS-based methods~\cite{poole2022dreamfusion, wang2023prolificdreamer} utilize a one-by-one optimization approach. \cref{tab:rp} reports the CLIP R-Precision~\citep{jun2023shap} and inference time on 50 test prompts manually derived from the captioned untrained Objaverse to quantitatively evaluate these methods. Also, optimization methods, Dreamfusion~\cite{poole2022dreamfusion} and ProlificDreamer~\citep{wang2023prolificdreamer}, are expensive, taking several hours to generate a single object. Moreover, these optimization methods may lead to more severe multi-face problems. In contrast, our method can produce realistic objects with reasonable geometry in only 40 seconds.
Furthermore, \model can serve as a strong prior for optimization-based methods and significantly boost their performance. Initializing the radiance field in ProlificDreamer~\cite{wang2023prolificdreamer} with our outputs shows remarkable improvements in both quality and computational efficiency, as shown in \cref{fig:comp_all}.

\begin{figure}
\setlength{\abovecaptionskip}{0cm}
\setlength{\belowcaptionskip}{-1em}
\centering{\includegraphics[width=0.48\textwidth]{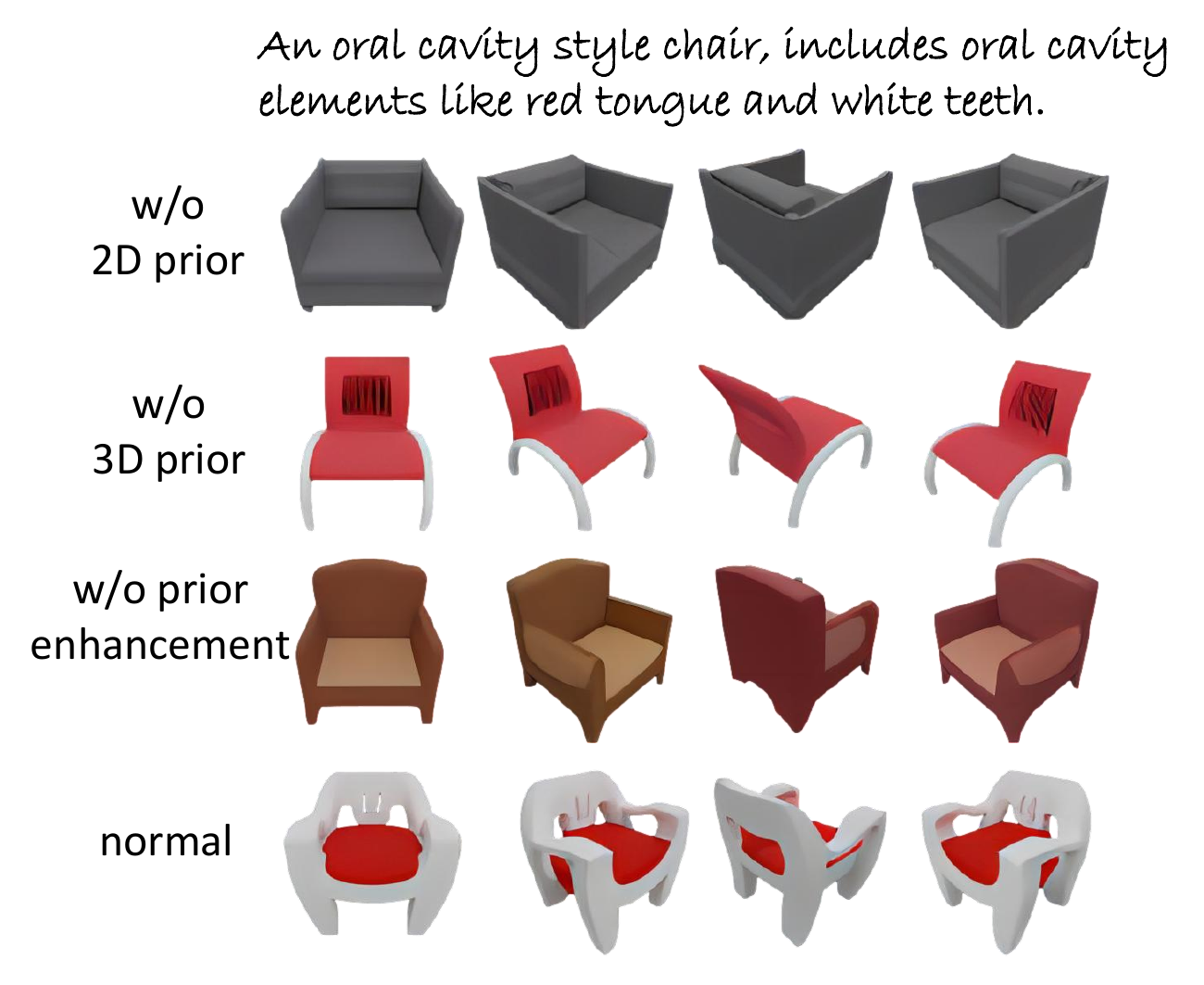
}}
\caption{Ablation of prior and prior enhancement.}
\vspace{-.5em}
\label{fig:ab}
\end{figure}

\begin{figure}
\setlength{\abovecaptionskip}{0cm}
\setlength{\belowcaptionskip}{-1.5em}
\centering{\includegraphics[width=0.45\textwidth]{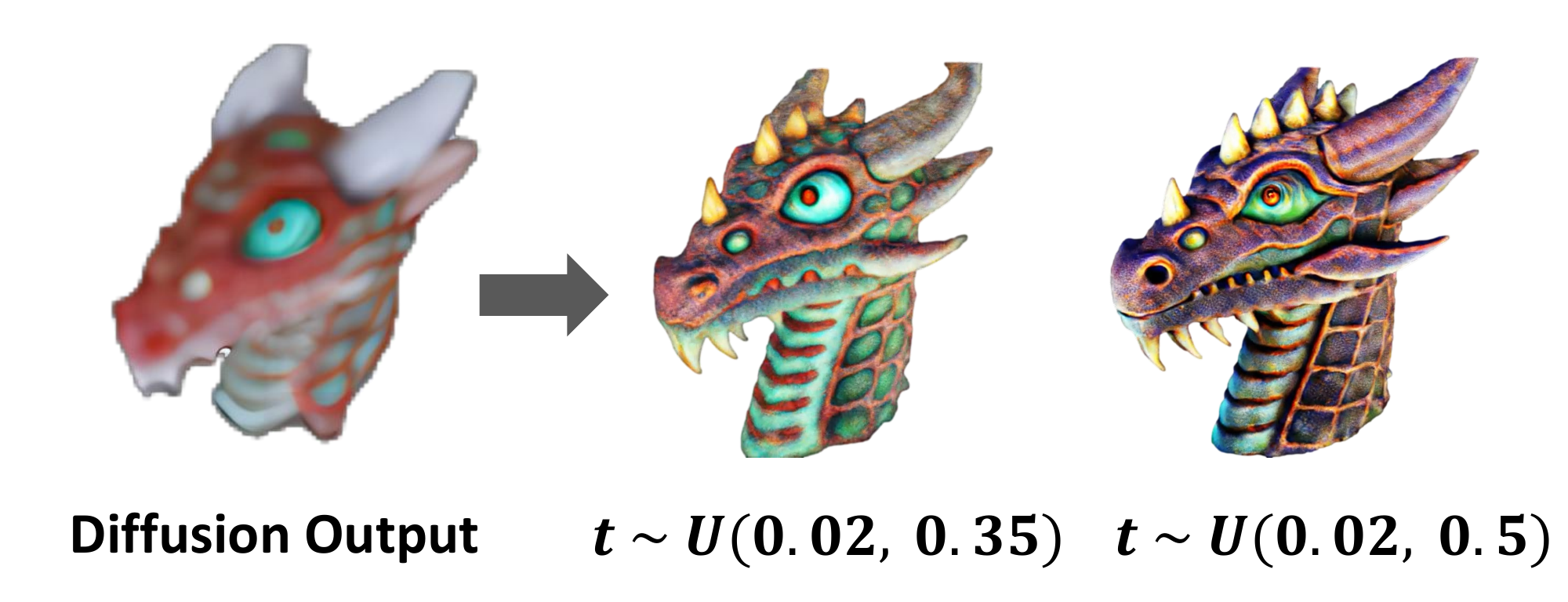}
}
\caption{Ablation of range of noise level $t$ for SDS.}
\label{fig:ab2}
\end{figure}

\vspace{-2mm}
\myparagraph{Comparison with multi-view methods}
Given one reference image, the multi-view method Zero-1-to-3~\citep{liu2023zero} produces images from novel viewpoints by fine-tuning a pre-trained 2D diffusion model on multi-view datasets. However, this method employs cross-view attention to establish multi-view correspondence without an inherent understanding of 3D structures, inevitably leading to inconsistent multi-view images as shown in \cref{fig:comp_all}. Moreover, the Zero-123 series cannot directly generate the 3D mesh, requiring substantial post-processing (SDS loss) to acquire the geometry. In contrast, our framework also incorporates 3D priors, in addition to 2D priors, and thus can generate more accurate 3D geometries. 

\subsection{Abalation Studies}
\label{sec:ab}
We perform comprehensive ablation studies on the ShapeNet-Chair dataset~\cite{chang2015shapenet} to evaluate the importance of each component below. More ablation results can be found in the supplementary material.
\vspace{-.7em}

\myparagraph{3D priors.}
\vspace{-.7em}
To assess the impact of 3D priors, we eliminate the conditional radiance field from Shap-E and train the 3D geometry generation from scratch. The experimental results in the second row of \cref{fig:ab} demonstrate that in the absence of the 3D priors, our framework can only generate common objects in the training set.
 
\vspace{-.7em}
\myparagraph{2D priors.}
\vspace{-.7em}
To delve into the impact of 2D priors, we randomly initiate the parameters of the 2D diffusion model, instead of fine-tuning on a pretrained model.
The results in the first row of \cref{fig:ab} show that in the absence of 2D priors, the textures generated tend to fit the stylistic attributes of the synthetic training data. Conversely, with 2D priors, we can produce more realistic textures.
\vspace{-.7em}
\myparagraph{Prior enhancement strategy.}
As discussed in \cref{sec:prior_enhance}, we can adjust the influence of both 3D and 2D priors by the prior enhancement strategy. \cref{fig:ab} also shows the results of not using this strategy. It shows that the prior enhancement strategy plays a vital role in achieving diverse and flexible 3D generation.
\vspace{-.7em}
\myparagraph{Range of noise level for SDS.}
The results in \cref{fig:ab2} illustrate the impact of the noise level during the entire optimization process, as discussed in \cref{sec:post_optim}. The 3D object generated with a smaller noise range is closer to the diffusion output. By adjusting the range of the noise level $t_{opt}$, we can effectively control the texture similarity between geometries before and after the optimization.
\vspace{-.3em}
\section{Conclusion}
In this paper, we propose Bidirectional Diffusion, which incorporates both 3D and 2D diffusion processes into a unified framework. Furthermore, Bidirectional Diffusion leverages the robust priors from 3D and 2D foundation models, achieving generalizable geometry and texture understanding. 
{
    \small
    \bibliographystyle{ieeenat_fullname}
    \bibliography{main}

\begin{thebibliography}{43}
\providecommand{\natexlab}[1]{#1}
\providecommand{\url}[1]{\texttt{#1}}
\expandafter\ifx\csname urlstyle\endcsname\relax
  \providecommand{\doi}[1]{doi: #1}\else
  \providecommand{\doi}{doi: \begingroup \urlstyle{rm}\Url}\fi

\bibitem[Achlioptas et~al.(2018)Achlioptas, Diamanti, Mitliagkas, and Guibas]{Achlioptas2018}
Panos Achlioptas, Olga Diamanti, Ioannis Mitliagkas, and Leonidas Guibas.
\newblock Learning representations and generative models for 3d point clouds.
\newblock In \emph{International conference on machine learning}, pages 40--49. PMLR, 2018.

\bibitem[Chang et~al.(2015)Chang, Funkhouser, Guibas, Hanrahan, Huang, Li, Savarese, Savva, Song, Su, et~al.]{chang2015shapenet}
Angel~X Chang, Thomas Funkhouser, Leonidas Guibas, Pat Hanrahan, Qixing Huang, Zimo Li, Silvio Savarese, Manolis Savva, Shuran Song, Hao Su, et~al.
\newblock Shapenet: An information-rich 3d model repository.
\newblock \emph{arXiv preprint arXiv:1512.03012}, 2015.

\bibitem[Chen and Zhang(2019)]{Chen2019}
Zhiqin Chen and Hao Zhang.
\newblock Learning implicit fields for generative shape modeling.
\newblock In \emph{Proceedings of IEEE Conference on Computer Vision and Pattern Recognition (CVPR)}, 2019.

\bibitem[Cheng et~al.(2022)Cheng, Lee, Tuyakov, Schwing, and Gui]{cheng2022sdfusion}
Yen-Chi Cheng, Hsin-Ying Lee, Sergey Tuyakov, Alex Schwing, and Liangyan Gui.
\newblock {SDFusion}: Multimodal 3d shape completion, reconstruction, and generation.
\newblock \emph{arXiv}, 2022.

\bibitem[Deitke et~al.(2022)Deitke, Schwenk, Salvador, Weihs, Michel, VanderBilt, Schmidt, Ehsani, Kembhavi, and Farhadi]{deitke2022objaverse}
Matt Deitke, Dustin Schwenk, Jordi Salvador, Luca Weihs, Oscar Michel, Eli VanderBilt, Ludwig Schmidt, Kiana Ehsani, Aniruddha Kembhavi, and Ali Farhadi.
\newblock Objaverse: A universe of annotated 3d objects.
\newblock \emph{arXiv preprint arXiv:2212.08051}, 2022.

\bibitem[Deng et~al.(2023)Deng, Jiang, Qi, Yan, Zhou, Guibas, and Anguelov]{deng2023nerdi}
C. Deng, C. Jiang, C.~R. Qi, X. Yan, Y. Zhou, L. Guibas, and D. Anguelov.
\newblock Nerdi: Single-view nerf synthesis with language-guided diffusion as general image priors.
\newblock In \emph{2023 IEEE/CVF Conference on Computer Vision and Pattern Recognition (CVPR)}, pages 20637--20647, 2023.

\bibitem[Erkoç et~al.(2023)Erkoç, Ma, Shan, Nießner, and Dai]{erkoç2023hyperdiffusion}
Ziya Erkoç, Fangchang Ma, Qi Shan, Matthias Nießner, and Angela Dai.
\newblock Hyperdiffusion: Generating implicit neural fields with weight-space diffusion, 2023.

\bibitem[Floyd(2023)]{deepfloyd_if}
Deep Floyd.
\newblock If project.
\newblock \url{https://github.com/deep-floyd/IF}, 2023.

\bibitem[Gao et~al.(2019)Gao, Yang, Wu, Yuan, Fu, Lai, , and Zhang]{Gao2019}
Lin Gao, Jie Yang, Tong Wu, Yu-Jie Yuan, Hongbo Fu, Yu-Kun Lai, , and Hao Zhang.
\newblock Sdm-net: Deep generative network for structured deformable mesh.
\newblock \emph{ACM Transactions on Graphics (TOG)}, 38:\penalty0 1--15, 2019.

\bibitem[Henzler et~al.(2019)Henzler, Mitra, and Ritschel]{Henzler2019}
Philipp Henzler, Niloy~J. Mitra, and Tobias Ritschel.
\newblock Escaping plato’s cave: 3d shape from adversarial rendering.
\newblock In \emph{The IEEE International Conference on Computer Vision (ICCV)}, 2019.

\bibitem[Ho and Salimans(2022)]{ho2022classifier}
Jonathan Ho and Tim Salimans.
\newblock Classifier-free diffusion guidance.
\newblock \emph{arXiv preprint arXiv:2207.12598}, 2022.

\bibitem[Ibing et~al.(2021)Ibing, Kobsik, and Kobbelt]{Ibing2021}
Moritz Ibing, Gregor Kobsik, and Leif Kobbelt.
\newblock Octree transformer: Autoregressive 3d shape generation on hierarchically structured sequences.
\newblock \emph{arXiv preprint arXiv:2111.12480}, 2021.

\bibitem[Jain et~al.(2022)Jain, Mildenhall, Barron, Abbeel, and Poole]{jain2022zero}
Ajay Jain, Ben Mildenhall, Jonathan~T Barron, Pieter Abbeel, and Ben Poole.
\newblock Zero-shot text-guided object generation with dream fields.
\newblock In \emph{Proceedings of the IEEE/CVF Conference on Computer Vision and Pattern Recognition}, pages 867--876, 2022.

\bibitem[Jun and Nichol(2023)]{jun2023shap}
Heewoo Jun and Alex Nichol.
\newblock Shap-e: Generating conditional 3d implicit functions.
\newblock \emph{arXiv preprint arXiv:2305.02463}, 2023.

\bibitem[Khalid et~al.(2022)Khalid, Xie, Belilovsky, and Tiberiu]{khalid2022clipmesh}
Nasir~Mohammad Khalid, Tianhao Xie, Eugene Belilovsky, and Popa Tiberiu.
\newblock Clip-mesh: Generating textured meshes from text using pretrained image-text models.
\newblock \emph{SIGGRAPH Asia 2022 Conference Papers}, 2022.

\bibitem[Li et~al.(2023)Li, Li, Savarese, and Hoi]{li2023blip}
Junnan Li, Dongxu Li, Silvio Savarese, and Steven Hoi.
\newblock Blip-2: Bootstrapping language-image pre-training with frozen image encoders and large language models.
\newblock \emph{arXiv preprint arXiv:2301.12597}, 2023.

\bibitem[Lin et~al.(2022)Lin, Gao, Tang, Takikawa, Zeng, Huang, Kreis, Fidler, Liu, and Lin]{lin2022magic3d}
Chen-Hsuan Lin, Jun Gao, Luming Tang, Towaki Takikawa, Xiaohui Zeng, Xun Huang, Karsten Kreis, Sanja Fidler, Ming-Yu Liu, and Tsung-Yi Lin.
\newblock Magic3d: High-resolution text-to-3d content creation.
\newblock \emph{arXiv preprint arXiv:2211.10440}, 2022.

\bibitem[Liu et~al.(2023{\natexlab{a}})Liu, Wu, Van~Hoorick, Tokmakov, Zakharov, and Vondrick]{liu2023zero}
Ruoshi Liu, Rundi Wu, Basile Van~Hoorick, Pavel Tokmakov, Sergey Zakharov, and Carl Vondrick.
\newblock Zero-1-to-3: Zero-shot one image to 3d object.
\newblock \emph{arXiv preprint arXiv:2303.11328}, 2023{\natexlab{a}}.

\bibitem[Liu et~al.(2023{\natexlab{b}})Liu, Feng, Black, Nowrouzezahrai, Paull, and Liu]{Liu2023MeshDiffusion}
Zhen Liu, Yao Feng, Michael~J. Black, Derek Nowrouzezahrai, Liam Paull, and Weiyang Liu.
\newblock Meshdiffusion: Score-based generative 3d mesh modeling.
\newblock In \emph{International Conference on Learning Representations}, 2023{\natexlab{b}}.

\bibitem[Long et~al.(2022)Long, Lin, Wang, Komura, and Wang]{long2022sparseneus}
Xiaoxiao Long, Cheng Lin, Peng Wang, Taku Komura, and Wenping Wang.
\newblock Sparseneus: Fast generalizable neural surface reconstruction from sparse views.
\newblock In \emph{European Conference on Computer Vision}, pages 210--227. Springer, 2022.

\bibitem[Melas-Kyriazi et~al.(2023)Melas-Kyriazi, Rupprecht, Laina, and Vedaldi]{melaskyriazi2023realfusion}
Luke Melas-Kyriazi, Christian Rupprecht, Iro Laina, and Andrea Vedaldi.
\newblock Realfusion: 360 reconstruction of any object from a single image.
\newblock In \emph{CVPR}, 2023.

\bibitem[Metzer et~al.(2022)Metzer, Richardson, Patashnik, Giryes, and Cohen-Or]{metzer2022latent}
Gal Metzer, Elad Richardson, Or Patashnik, Raja Giryes, and Daniel Cohen-Or.
\newblock Latent-nerf for shape-guided generation of 3d shapes and textures.
\newblock \emph{arXiv preprint arXiv:2211.07600}, 2022.

\bibitem[Metzer et~al.(2023)Metzer, Richardson, Patashnik, Giryes, and Cohen-Or]{metzer2023latent}
Gal Metzer, Elad Richardson, Or Patashnik, Raja Giryes, and Daniel Cohen-Or.
\newblock Latent-nerf for shape-guided generation of 3d shapes and textures.
\newblock In \emph{Proceedings of the IEEE/CVF Conference on Computer Vision and Pattern Recognition}, pages 12663--12673, 2023.

\bibitem[M\"uller et~al.(2022)M\"uller, Evans, Schied, and Keller]{mueller2022instant}
Thomas M\"uller, Alex Evans, Christoph Schied, and Alexander Keller.
\newblock Instant neural graphics primitives with a multiresolution hash encoding.
\newblock \emph{ACM Trans. Graph.}, 41\penalty0 (4):\penalty0 102:1--102:15, 2022.

\bibitem[Nichol et~al.(2022)Nichol, Jun, Dhariwal, Mishkin, and Chen]{nichol2022point}
Alex Nichol, Heewoo Jun, Prafulla Dhariwal, Pamela Mishkin, and Mark Chen.
\newblock Point-e: A system for generating 3d point clouds from complex prompts.
\newblock \emph{arXiv preprint arXiv:2212.08751}, 2022.

\bibitem[Park et~al.(2019)Park, Florence, Straub, Newcombe, , and Lovegrove]{Park2019}
Jeong~Joon Park, Peter Florence, Julian Straub, Richard Newcombe, , and Steven Lovegrove.
\newblock Deepsdf: Learning continuous signed distance functions for shape representation.
\newblock In \emph{Proceedings of IEEE Conference on Computer Vision and Pattern Recognition (CVPR)}, pages 165--174, 2019.

\bibitem[Poole et~al.(2022)Poole, Jain, Barron, and Mildenhall]{poole2022dreamfusion}
Ben Poole, Ajay Jain, Jonathan~T Barron, and Ben Mildenhall.
\newblock Dreamfusion: Text-to-3d using 2d diffusion.
\newblock \emph{arXiv preprint arXiv:2209.14988}, 2022.

\bibitem[Qian et~al.(2023)Qian, Mai, Hamdi, Ren, Siarohin, Li, Lee, Skorokhodov, Wonka, Tulyakov, et~al.]{qian2023magic123}
Guocheng Qian, Jinjie Mai, Abdullah Hamdi, Jian Ren, Aliaksandr Siarohin, Bing Li, Hsin-Ying Lee, Ivan Skorokhodov, Peter Wonka, Sergey Tulyakov, et~al.
\newblock Magic123: One image to high-quality 3d object generation using both 2d and 3d diffusion priors.
\newblock \emph{arXiv preprint arXiv:2306.17843}, 2023.

\bibitem[Radford et~al.(2021)Radford, Kim, Hallacy, Ramesh, Goh, Agarwal, Sastry, Askell, Mishkin, Clark, et~al.]{radford2021clip}
Alec Radford, Jong~Wook Kim, Chris Hallacy, Aditya Ramesh, Gabriel Goh, Sandhini Agarwal, Girish Sastry, Amanda Askell, Pamela Mishkin, Jack Clark, et~al.
\newblock Learning transferable visual models from natural language supervision.
\newblock In \emph{International conference on machine learning}, pages 8748--8763. PMLR, 2021.

\bibitem[Rombach et~al.(2022)Rombach, Blattmann, Lorenz, Esser, and Ommer]{rombach2022high}
Robin Rombach, Andreas Blattmann, Dominik Lorenz, Patrick Esser, and Bj{\"o}rn Ommer.
\newblock High-resolution image synthesis with latent diffusion models.
\newblock In \emph{Proceedings of the IEEE/CVF Conference on Computer Vision and Pattern Recognition}, pages 10684--10695, 2022.

\bibitem[Saharia et~al.(2022)Saharia, Chan, Saxena, Li, Whang, Denton, Ghasemipour, Gontijo~Lopes, Karagol~Ayan, Salimans, et~al.]{saharia2022photorealistic}
Chitwan Saharia, William Chan, Saurabh Saxena, Lala Li, Jay Whang, Emily~L Denton, Kamyar Ghasemipour, Raphael Gontijo~Lopes, Burcu Karagol~Ayan, Tim Salimans, et~al.
\newblock Photorealistic text-to-image diffusion models with deep language understanding.
\newblock \emph{Advances in Neural Information Processing Systems}, 35:\penalty0 36479--36494, 2022.

\bibitem[Seo et~al.(2023)Seo, Jang, Kwak, Ko, Kim, Kim, Kim, Lee, and Kim]{seo2023let}
Junyoung Seo, Wooseok Jang, Min-Seop Kwak, Jaehoon Ko, Hyeonsu Kim, Junho Kim, Jin-Hwa Kim, Jiyoung Lee, and Seungryong Kim.
\newblock Let 2d diffusion model know 3d-consistency for robust text-to-3d generation.
\newblock \emph{arXiv preprint arXiv:2303.07937}, 2023.

\bibitem[Shi et~al.(2023)Shi, Wang, Ye, Mai, Li, and Yang]{shi2023MVDream}
Yichun Shi, Peng Wang, Jianglong Ye, Long Mai, Kejie Li, and Xiao Yang.
\newblock Mvdream: Multi-view diffusion for 3d generation.
\newblock \emph{arXiv:2308.16512}, 2023.

\bibitem[Smith and Meger(2017)]{Edward2017}
Edward Smith and David Meger.
\newblock Deep unsupervised learning using nonequilibrium thermodynamics.
\newblock In \emph{Conference on Robot Learning}, pages 87--96. PMLR, 2017.

\bibitem[Wang et~al.(2022)Wang, Du, Li, Yeh, and Shakhnarovich]{wang2022score}
Haochen Wang, Xiaodan Du, Jiahao Li, Raymond~A Yeh, and Greg Shakhnarovich.
\newblock Score jacobian chaining: Lifting pretrained 2d diffusion models for 3d generation.
\newblock \emph{arXiv preprint arXiv:2212.00774}, 2022.

\bibitem[Wang et~al.(2021)Wang, Liu, Liu, Theobalt, Komura, and Wang]{wang2021neus}
Peng Wang, Lingjie Liu, Yuan Liu, Christian Theobalt, Taku Komura, and Wenping Wang.
\newblock Neus: Learning neural implicit surfaces by volume rendering for multi-view reconstruction.
\newblock \emph{arXiv preprint arXiv:2106.10689}, 2021.

\bibitem[Wang et~al.(2023)Wang, Lu, Wang, Bao, Li, Su, and Zhu]{wang2023prolificdreamer}
Zhengyi Wang, Cheng Lu, Yikai Wang, Fan Bao, Chongxuan Li, Hang Su, and Jun Zhu.
\newblock Prolificdreamer: High-fidelity and diverse text-to-3d generation with variational score distillation.
\newblock \emph{arXiv preprint arXiv:2305.16213}, 2023.

\bibitem[Wu et~al.(2016)Wu, Zhang, Xue, Freeman, and Tenenbaum]{Wu2016}
Jiajun Wu, Chengkai Zhang, Tianfan Xue, Bill Freeman, and Josh Tenenbaum.
\newblock Learning a probabilistic latent space of object shapes via 3d generative-adversarial modeling.
\newblock In \emph{Advances in neural information processing systems}, pages 82--90, 2016.

\bibitem[Wu et~al.(2023)Wu, Zhang, Fu, Wang, Ren, Pan, Wu, Yang, Jiaqi~Wang, Lin, and Liu]{wu2023omniobject3d}
Tong Wu, Jiarui Zhang, Xiao Fu, Yuxin Wang, Jiawei Ren, Liang Pan, Wayne Wu, Lei Yang, Chen~Qian Jiaqi~Wang, Dahua Lin, and Ziwei Liu.
\newblock Omniobject3d: Large-vocabulary 3d object dataset for realistic perception, reconstruction and generation.
\newblock \emph{IEEE/CVF Conference on Computer Vision and Pattern Recognition (CVPR)}, 2023.

\bibitem[Yang et~al.(2019)Yang, Huang, Hao, Liu, Belongie, and Hariharan]{yang2019pointflow}
Guandao Yang, Xun Huang, Zekun Hao, Ming-Yu Liu, Serge Belongie, and Bharath Hariharan.
\newblock Pointflow: 3d point cloud generation with continuous normalizing flows.
\newblock In \emph{Proceedings of the IEEE/CVF international conference on computer vision}, pages 4541--4550, 2019.

\bibitem[Yu et~al.(2021)Yu, Ye, Tancik, and Kanazawa]{yu2021pixelnerf}
Alex Yu, Vickie Ye, Matthew Tancik, and Angjoo Kanazawa.
\newblock pixelnerf: Neural radiance fields from one or few images.
\newblock In \emph{Proceedings of the IEEE/CVF Conference on Computer Vision and Pattern Recognition}, pages 4578--4587, 2021.

\bibitem[Zeng et~al.(2022)Zeng, Vahdat, Williams, Gojcic, Litany, Fidler, and Kreis]{zeng2022lion}
Xiaohui Zeng, Arash Vahdat, Francis Williams, Zan Gojcic, Or Litany, Sanja Fidler, and Karsten Kreis.
\newblock Lion: Latent point diffusion models for 3d shape generation.
\newblock \emph{arXiv preprint arXiv:2210.06978}, 2022.

\bibitem[Zhang and Agrawala(2023)]{zhang2023controlnet}
Lvmin Zhang and Maneesh Agrawala.
\newblock Adding conditional control to text-to-image diffusion models.
\newblock \emph{arXiv preprint arXiv:2302.05543}, 2023.

\end{thebibliography}
}

\clearpage
\setcounter{page}{1}
\maketitlesupplementary

In the supplementary material, we first introduce the data processing pipeline in (\S ~\ref{sec:data-pre}), then provide more implementation details of the model architecture (\S ~\ref{sec:model-detail}), more training details in (\S ~\ref{sec:train-detail}), and give more ablation results in (\S ~\ref{sec:more-results}).

\subsection{Data Processing}
\label{sec:data-pre}
As mentioned in the main paper, we use 6k ShapNet-Chair~\cite{chang2015shapenet}
and LVIS Objaverse 40k ~\cite{deitke2022objaverse} as our training datasets. We obtain the Objaverse 40k dataset by filtering objects with LVIS category labels in the 800k Objaverse data. To process data for the 2D diffusion process, we use Blender to render each 3D object into 8 images with a fixed elevation of $30^\circ$ and evenly distributed azimuth from $-180^\circ$ to $180^\circ$. These fixed view images serve as the ground truth multi-view image set $\mathcal{V}$. In addition, we also randomly render 16 views to supervise the novel view rendering of the denoised radiance field $\mathcal{F}_{0}'$. All the images are rendered at a resolution of $256 \times 256$. Since we adopt the DeepFloyd as our 2D foundation model which runs at a resolution of $64 \times 64$, the rendered images are downsampled to $64 \times 64$ during training. To process data for the 3D diffusion, we compute the signed distance of each 3D object at each $N \times N \times N$ grid point within a $[-1, 1]$ cube, where $N$ is set to 128 in our experiments. To obtain the latent code $\mathcal{C}$ for each object, we use the encoder in Shap-E~\cite{jun2023shap} to encode each object and apply $t_0=0.4$ level Gaussian noise to $\mathcal{C}$ to get noisy $\mathcal{C}_{t_0}$, and then decode the condition radiance field during training.

Furthermore, both the ShapNet-Chair and Objaverse dataset contains no text prompts, so we use Blip-2~\cite{li2023blip} to generate labels for the Objaverse object by rendering the image from a positive view. For evaluation, we manually choose 50 text prompts from the Objaverse dataset without LVIS label, ensuring the text prompts have not been trained during training.

\subsection{Model Architecture Details}
\label{sec:model-detail}
Our framework contains a 3D denoising network built upon 3D SparseConv U-Net and a 2D denoising network built upon 2D U-Net. Below we provide more details for each of them.

\begin{figure*}[htbp]
\setlength{\belowcaptionskip}{-0.3cm}
\centering{\includegraphics[width=0.85\linewidth]{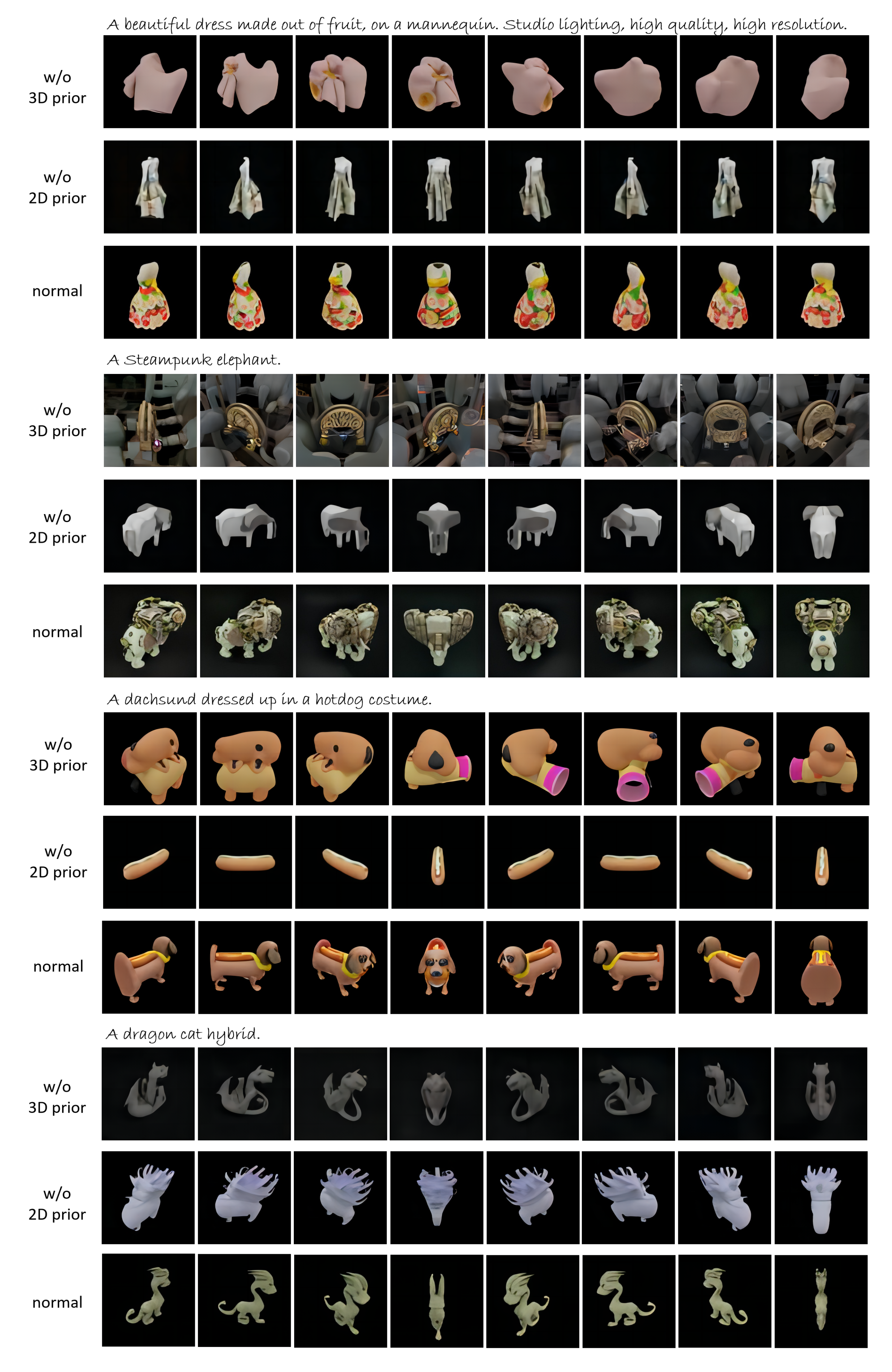}}
\vspace{-1em}
\caption{More ablation results showing the importance of both 2D and 3D priors in our model.}
\label{fig:more_ab}
\end{figure*}

\subsubsection{3D Denoising Network}
Given the input feature volume 
\begin{equation}
\begin{split}
\mathcal{S}_{\text{in}} = \text{Concat}(&\mathcal{M}, \text{Sp3DConv}(\mathcal{N}), \\
&\text{Sp3DConv}(\mathcal{G}_{t_0}))
\end{split}
\end{equation}
as discussed in Section 3.2 of the main paper, we use a 3D sparse U-Net $\mathcal{U}$ to denoise the signed distance field. Specifically, we first use a $1 \times 1 \times 1$ convolution to adjust the number of input channels to 128. Then we stack four $3 \times 3 \times 3$ sparse 3D convolution blocks to extract hierarchical features while obtaining downsampled $8 \times 8 \times 8$ feature grids. It is noteworthy that we inject the timestep and text embeddings into each sparse convolution block to make the network aware of the current noise level and text information. In practice, we first use an MLP to project the scalar timestep $t$ to high-dimensional features and fuse it with the text embeddings with another MLP to get the fused embeddings as follows:
\begin{equation}
    \text{emb} = \text{MLP}_{2}(\text{Concat}(\text{emb}_{\text{text}}, \text{MLP}_{1}(t))), 
\end{equation}
where $\text{emb}_{\text{text}}$ denotes the text embeddings. Then in each sparse convolution block, we project the fused embeddings to scale $\beta$ and shift $\gamma$:
\begin{equation}
    \beta, \gamma = \text{Chunk}(\text{MLP}_{\text{proj}}(\text{GeLU}(\text{emb}))),
\end{equation}
where \text{GeLU} is activated function, Chunk operation splits the projected features into two equal parts along the channel dimension.
After that, we introduce modulation to the sparse convolution by:
\begin{equation}
    \mathcal{S}_{k+1} = (1 + \beta) (\text{SparseConv}(\text{GroupNorm}(\mathcal{S}_{k}))) + \gamma,
\end{equation}
where k denotes the feature level, $\mathcal{S}_{k}$ and $\mathcal{S}_{k+1}$ are the input and output of the $k$-th level sparse convolution block.
Subsequently, we use 4 sparse deconvolution blocks to upsample the bottleneck feature grids with residuals linked from the extracted hierarchical features:
\begin{equation}
    \mathcal{S}_{k}' = \text{SparseDeConv}(\mathcal{S}_{k+1}') + \mathcal{S}_{k},
\end{equation}
where $\mathcal{S}'_{k+1}$ and $\mathcal{S}'_{k}$ are the input and output of the $k$-th level sparse de-convolution block,
and obtain the output features $\mathcal{S}$ of the 3D U-Net.

To obtain the denoised signed distance field, we first query each 3D position $p$ in the fused feature grid $\mathcal{S}$ to fetch its feature $\mathcal{S}(p)$ by Trilinear Interpolation. Then we apply several MLPs (we adopt the ResNetFC blocks in ~\cite{yu2021pixelnerf}) to predict the signed distance at position $p$:
\begin{equation}
    \mathcal{F}_{0}' = \text{MLP}(\mathcal{S}(p), \lambda(p)),
\end{equation}
where $\lambda(p)$ is the positional encoding:
\begin{equation}
\begin{split}
    \lambda(p) = &(\text{sin}(2^0\omega p), \text{cos}(2^0\omega p), \text{sin}(2^1\omega p), \text{cos}(2^1\omega p),  \\
    &..., \text{sin}(2^{L-1}\omega p), \text{cos}(2^{L-1}\omega p)).
\end{split}
\end{equation}
$L$ is set to 6 in all experiments.

\begin{figure*}[htbp]
\setlength{\belowcaptionskip}{-0.3cm}
\centering{\includegraphics[width=0.98\linewidth]{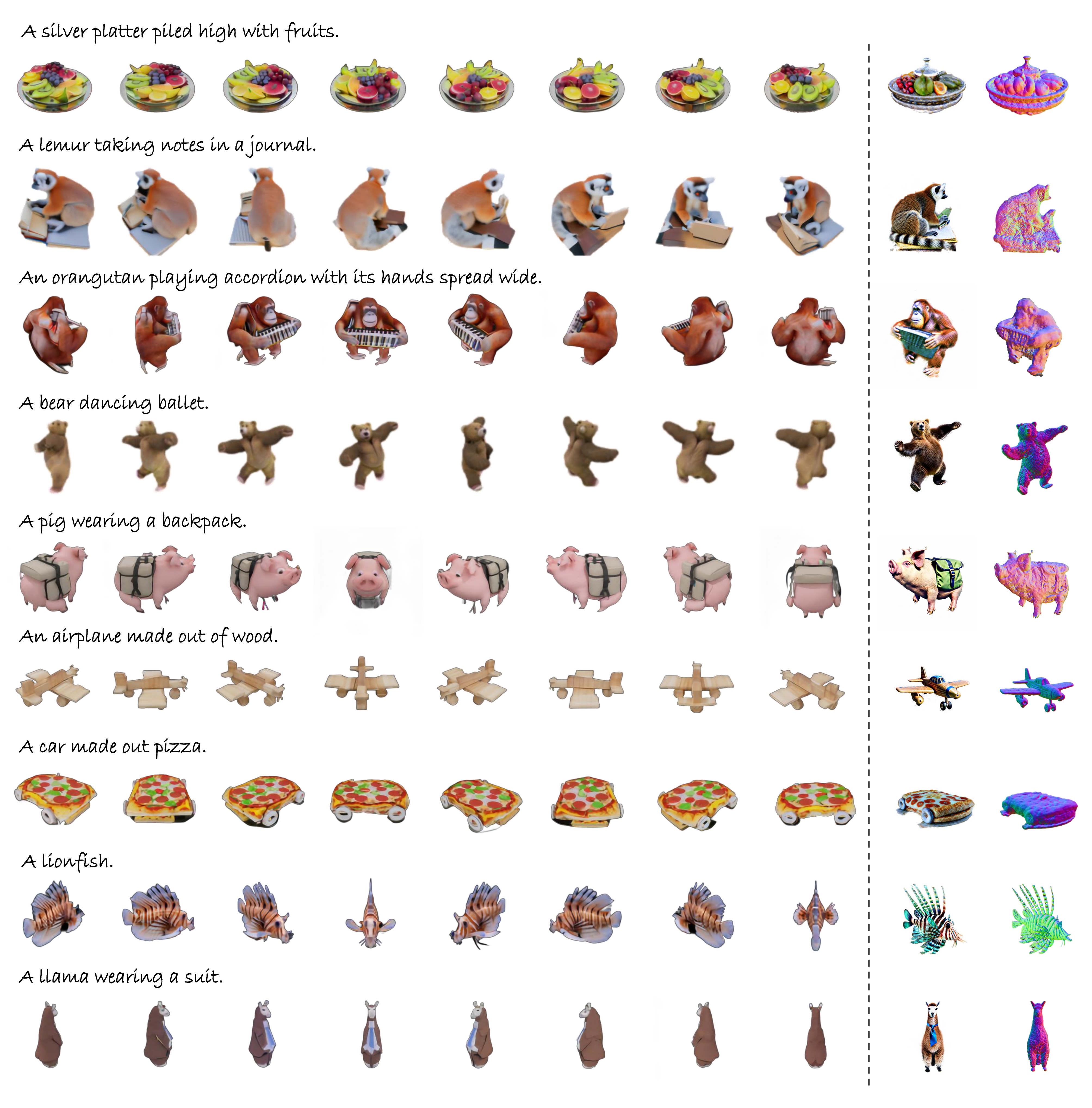}}
\vspace{-1em}
\caption{More generated 3D objects by our model. Left side shows the diffusion output and right side shows the 3D object after optimization.}
\label{fig:more_res}
\end{figure*}

\subsubsection{2D Denoising Network}
Our 2D denoising network contains a U-Net of the 2D foundation model (DeepFloyd) and a ControlNet~\cite{zhang2023controlnet} modulation module to jointly denoise the multi-view image set. In practice, given the $M$ noisy images $\mathcal{V}_t = \left \{ \mathcal{I}_t^i \right \}_{i=1}^{M}$ from the 2D diffusion process and $M$ rendered images $\left \{ \mathcal{H}^{i} \right \}_{i=1}^{M}$ from the 3D diffusion process as mentioned in Section 3.3 of the main paper, we first reshape both of them from $[B, M, C, H, W]$ to $[B \times M, C, H, W]$, where $B, C, H, W$ denote batch size, channel, height, width, respectively. 
Then we feed the noisy images to the frozen encoder $\mathcal{E}^*$ of DeepFloyd to get encoded features:
\begin{equation}
    P = \mathcal{E}^{*}(\text{Reshape}(\left \{ \mathcal{I}_{t}^{i} \right \}_{i=1}^{M}), t, \text{emb}_{\text{text}}).
\end{equation}
$P = \left \{ p^{k} \right \}_{k=1}^{K}$ where $p^{k}$ denotes the $k$-th features of the total $K$ feature levels.
Simultaneously, we feed the rendered images to the trainable copy encoder $\mathcal{E}$ of ControlNet to obtain the hierarchical 3D consistent condition features:
\begin{equation}
    Q = \mathcal{E}(\text{Reshape}(\left \{ \mathcal{H}^{i} \right \}_{i=1}^{M}), t, \text{emb}_{\text{text}}),
\end{equation}
where $Q = \left \{ q^{k} \right \}_{k=1}^{K}$. Subsequently, we decode $P$ with the frozen decoder $\mathcal{D}^{*}$ of DeepFloyd and the condition residual features $Q$. Specifically, in the $k$-th decoding stage, we first apply zero-convolutions to the condition feature $q^{k}$ and then add it to the original decoded features as residuals:
\begin{equation}
    \hat{f}^k = p^{k} + \mathcal{D}_{k-1}^{*}(p^{k-1}) + \text{ZeroConv}(q^{k}),
\end{equation} where $\mathcal{D}_{k-1}^{*}$ denotes the $k-1$-th frozen decoding layer of DeepFloyd. In this way, we can denoise the multi-view noisy images in a unified manner by introducing the 3D consistent condition signal as guidance. In practice, we set $M=8$ in our experiments.

\begin{figure}[htbp]
\setlength{\belowcaptionskip}{-0.3cm}
\centering{\includegraphics[width=0.98\linewidth]{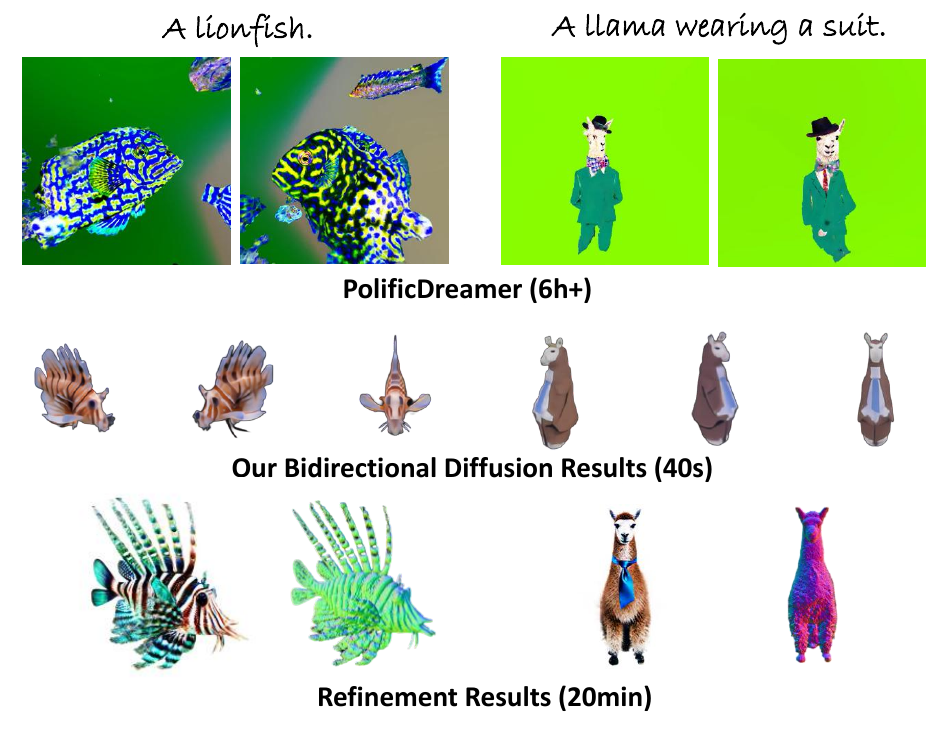}}
\vspace{-1em}
\caption{Comparison between our results with the object directly generated by the optimization method (ProlificDreamer).}
\label{fig:more_comp}
\end{figure}

\subsection{More Training Details}
\label{sec:train-detail}
We train our framework on 4 NVIDIA A100 GPUs with a batch size of 4. For ShapeNet-Chair, the training takes about 8 hours to converge. For Objaverse 40k, the training takes 5 days. We use the AdamW optimizer with $\beta = (0.9, 0.999)$ and weight decay $=0.01$. Notably, we set the learning rate of the 2D diffusion model to $2 \times 10^{-6}$ while using a much larger learning rate of $5 \times 10^{-5}$ for the 3D diffusion model.

\begin{figure}[htbp]
\setlength{\belowcaptionskip}{-0.3cm}
\centering{\includegraphics[width=0.98\linewidth]{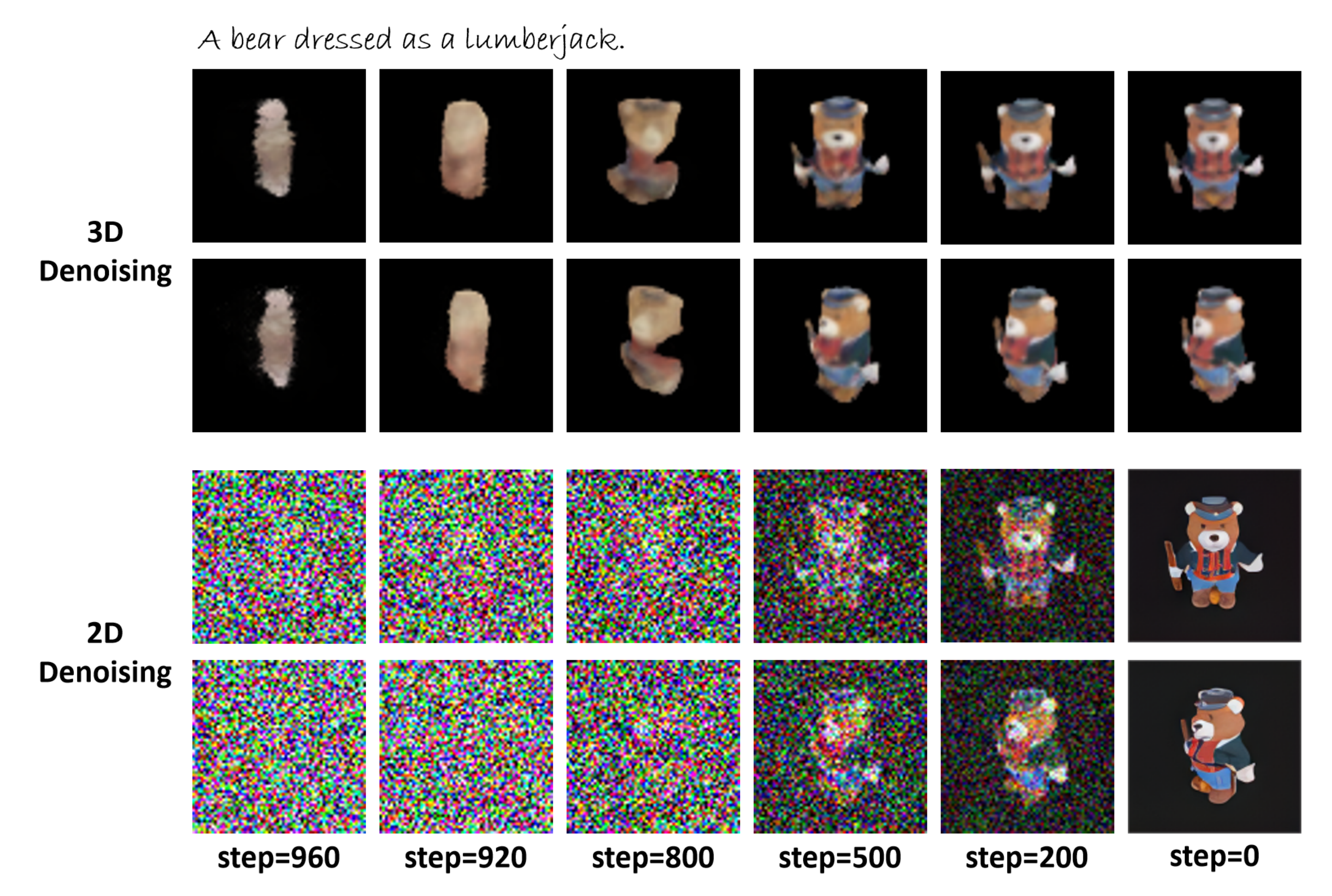}}
\vspace{-1em}
\caption{Visualization of our 2D and 3D denoising processes (the maximum diffusion step is 1,000). The top two rows show the rendering views of the implicit field during the 3D denoising process, and the bottom two rows show the 2D sample results during the 2D denoising process.}
\label{fig:vis_denoise}
\end{figure}

\subsection{More Experiments}
\label{sec:more-results}

\subsubsection{Ablation for Priors}
In \cref{fig:more_ab}, we provide additional results for the ablation of 3D and 2D priors mentioned in \cref{sec:ab}. Our method can produce more realistic textures with 2D priors and more robust geometry with 3D priors.

\subsubsection{Visualization of 2D-3D Denoising}
We also demonstrated the visualization of 2D and 3D denoising processes during bidirectional diffusion sampling as shown in \cref{fig:vis_denoise}. The top two lines show the rendering views of the implicit field during the 3D denoising process, and the bottom two lines show the 2D sample results during the 2D denoising process. 3D and 2D representations are jointly denoised, and in the early step of diffusion sampling, 3D representations can provide basic geometric shapes, which guides 2D diffusion to generate geometrically reasonable images. In the later step of sampling, texture generation is dominated by 2D diffusion.

\subsubsection{More Results}
In \cref{fig:more_res}, we provide more high-quality results generated by our entire framework. And in \cref{fig:more_comp}, we demonstrated a comparison with the previous state-of-the-art optimization method~\cite{wang2023prolificdreamer}]. Our approach not only significantly reduces time costs but is also more robust in understanding geometry.

\end{document}